\documentclass[3p,12pt,hidelinks, fleqn]{elsarticle}
\usepackage{amsmath, amssymb}
 \usepackage[fleqn]{nccmath}
\usepackage{cases} 
\usepackage{lineno,hyperref}
\usepackage{cleveref}
\usepackage{cases} 
\usepackage[linesnumbered, ruled,vlined]{algorithm2e}
\usepackage{algorithmic}
\usepackage{capt-of}
\usepackage{soul}
\usepackage{flushend} 
\usepackage[tight,footnotesize]{subfigure}
\usepackage[font=small,labelfont=bf,tableposition=top]{caption}
\usepackage{float}
\usepackage{array}
\usepackage{subfiles}
\usepackage{graphicx,subfigure}
\usepackage{multirow}
\usepackage[table,xcdraw]{xcolor}
 \usepackage{tabularx,booktabs, threeparttable, caption}%
\usepackage{longtable,tabu}
\SetAlFnt{\footnotesize}
\SetKw{KwDownTo}{downto}
\SetKw{KwTrue}{true}
\SetKw{KwFalse}{false}
\SetKwInOut{Input}{Input}
\SetKwInOut{Output}{Output}
\SetKw{KwAnd}{and}
\usepackage{pgfplots}
\usepackage{comment}
\usepackage{amsthm}
\usepackage{rotating}
\usepackage{placeins}
\modulolinenumbers[5]
\newcommand{\subsubsubsection}[1]{\paragraph{#1}\mbox{}\\}
\usepackage{mathtools}
\DeclarePairedDelimiter{\ceil}{\lceil}{\rceil}
\DeclarePairedDelimiter\floor{\lfloor}{\rfloor}

\journal{Journal of \LaTeX\ Templates}
\makeatletter
\def\ps@pprintTitle{%
  \let\@oddhead\@empty
  \let\@evenhead\@empty
  \let\@oddfoot\@empty
  \let\@evenfoot\@oddfoot
}
\makeatother

\usepackage[acronym]{glossaries} 
\makeglossaries
\newacronym{wsn}{WSN}{Wireless Sensor Network}
\newacronym{wrsn}{WRSN}{Wireless Rechargeable Sensor Network}
\newacronym{ga}{GA}{Genetic Algorithm}
\newacronym{mfea}{MFEA}{Multifactorial Evolutionary Algorithm}
\newacronym{iot}{IoT}{Internet of Thing}
\newacronym{mc}{MC}{Mobile Charger}
\newacronym{bs}{BS}{Base Station}
\newacronym{edmp}{EDMP}{ Energy Depletion Minimization Problem}
\newacronym{cmaes}{CMAES}{Covariance Matrix Adaptation Evolution Strategy}

\begin{document}

\begin{frontmatter}
\title{Minimizing the energy depletion in wireless rechargeable sensor networks using bi-level metaheuristic charging schemes}

\author[hust]{Huynh Thi Thanh Binh\corref{cor1}}
\ead{binhht@soict.hust.edu.vn}

\author[hust]{Le Van Cuong}
\ead{cuonglv.hust@gmail.com}

\author[hust]{Dang Hai Dang}
\ead{dang.dh224944@sis.hust.edu.vn}

\author[hus]{Le Trong Vinh}
\ead{vinhlt@vnu.edu.vn}

\cortext[cor1]{Corresponding author.}

\address[hust]{School of Information and Communication Technology, Hanoi University of Science and Technology, Vietnam}
\address[hus]{University of Science, Vietnam National University, Hanoi}

\begin{abstract}
Recently, Wireless Rechargeable Sensor Networks (WRSNs) that leveraged the advantage of wireless energy transfer technology have opened a promising opportunity in solving the limited energy issue. However, an ineffective charging strategy may reduce the charging performance. Although many practical charging algorithms have been introduced, these studies mainly focus on optimizing the charging path with a fully charging approach. This approach may lead to the death of a series of sensors due to their extended charging latency. This paper introduces a novel partial charging approach that follows a bi-level optimized scheme to minimize energy depletion in WRSNs. We aim at optimizing simultaneously two factors: the charging path and time. To accomplish this, we first formulate a mathematical model of the investigated problem. We then propose two approximate algorithms in which the optimization of the charging path and the charging time are considered as the upper and lower level, respectively. The first algorithm combines a Multi-start Local Search method and a Genetic Algorithm to find a solution. The second algorithm adopts a nested approach that utilizes the advantages of the Multitasking and Covariance Matrix Adaptation Evolutionary Strategies. Experimental validations on various network scenarios demonstrate that our proposed algorithms outperform the existing works.

\end{abstract}
  

\end{frontmatter}

\section{Introduction}
A \gls{wsn} consists of 
a collection of battery-powered sensor nodes deployed in a region of interest to monitor the physical environment and transfer the sensing information to the \gls{bs} via multi-hop communication. \cite{kocakulak2017overview,le2018load,kandris2020applications}. Nowadays, \glspl{wsn} play a nucleus role in the \glspl{iot} revolution due to their wide applications from civilian to military such as the smart metropolis, environment monitoring, intrusion detection, and battlefield surveillance \cite{ noel2017structural,le2019telpac, min2012multi}. 
However, limited energy issues remain as a major bottleneck phenomenon in \glspl{wsn}. When a sensor's battery is exhausted, the sensor becomes a dead node and loses its monitoring and communicating ability causing a series of negative impacts on the whole network performance \cite{kocakulak2017overview,  huong2020genetic}. 
Therefore, one of the most critical conditions in continuously maintaining the network's operation is to avoid the energy depletion of the sensor nodes. Energy-saving methods have been applied to prolong the sensor lifetime during the past decade \cite{le2018load, le2019node}. However, these approaches can only extend the network lifetime for a limited period. In recent years, a new approach that exploits the strength of wireless energy transfer technology has opened a promising solution for the energy issue in \glspl{wsn}. Specifically, this technology allows energy transmitting from one or multiple \glspl{mc} to the sensor nodes (equipped with a wireless energy receiver) without any wires or plugins. As a result, it has carried out a new network generation, named ``\gls{wrsn}''.  
In \gls{wrsn}, the \gls{mc} periodically travels around the network and charges sensors following either the on-demand charging strategy \cite{xu2018maximizing, feng2016starvation, kaswan2018efficient, zhu2018adaptive, long2023q}, or the periodic charging strategy \cite{ huong2020genetic, fu2015optimal, ouyang2020utility}. Regarding the first strategy, the MC will travels and charges the requests receive from sensor nodes having its remaining energy below a predefined threshold. Consider the second strategy, the \gls{mc} travels following a predetermined path to charge sensor nodes. 
In \glspl{wrsn}, as the charging scheme plays a decisive role in prolonging the network lifetime, charging algorithm optimization thus has become a challenging problem. 

Although many charging schemes have been proposed, most of the current works focus on optimizing the charging path of the \gls{mc} (i.e., a sequence of the charging locations sorted by the visiting order of the \gls{mc}). The studies \cite{kaswan2018efficient, xie2014multi,fu2015esync, ye2017charging, xu2015efficient, lyu2019periodic} aim at finding the optimal charging path with the fully charging method where the sensor's battery is charged to maximum capacity. Moreover, they assumed that the \gls{mc} energy is unlimited or sufficient enough to charge all sensor nodes in each charging round.
The full charging method may result in a long charging session of the MC if the number of sensors is large. It is only suitable for small networks with an insignificant number of sensor nodes. 
Besides, the charging time (i.e., the period that the \gls{mc} stays and charges at each charging location) is also a primary factor in deciding the sensor's lifetime. If the \gls{mc} spends too much time at a charging location, the waiting time of other sensors will be increased; thus, uncharged sensors' energy can be exhausted before being charged. On the other hand, if the \gls{mc} allocates too little charging time for each sensor node, it will not have enough energy to operate until the next charging cycle. 

Recently, the works in \cite{xu2016maximizing,nguyen2020extending, carrabs2020optimization} attempted to solve the problem of optimizing charging time at each charging position to maximize the network lifetime. The authors \cite{nguyen2020extending} introduced a heuristic algorithm based on the sensor's parameters
to estimate the charging time interval at each location. The authors \cite{carrabs2020optimization} addressed the problem of deciding the charging amount for each sensor node to maximize the target covering time. The above-mentioned studies assume that the MC's charging path is given and fixed in every charging round. 
The study \cite{huong2020genetic} was the first to investigate the problem of minimizing the number of energy-depleted sensors by jointly optimizing both the charging path and the charging time. They decomposed the original problem into two sub-problems: charging path determining and charging time identifying, and then solved them separately. 
In addition, the previous periodic charging schemes assume that all sensors have the same energy consumption rate \cite{shi2011renewable, xu2015efficient, ouyang2020utility}. That is not realistic due to dynamic parameters in the network environment.

In this paper, we investigate a novel partial charging approach to minimize the number of dead sensor nodes under constraints in which the MC's battery is limited, and sensors' energy consumption rate is diverse. Specifically, we aim at optimizing both the charging path and charging time at each sensor node simultaneously. We named our investigated problem as Energy Depletion Minimization Problem (EDMP) in \glspl{wrsn}. To solve the problem, we propose two approximate algorithms that leverage the advantage of the bi-level optimization approach, where determining the optimal charging path and charging time are considered the upper level and lower level optimization tasks, respectively. 
 The first algorithm, named MLSGA based on a multi-restart mechanism to explore multiple feasible points in the enormous search space and escape the local optimum to find the global solution. A feasible solution of the charging scheme consists of a charging path and a charging time sequence correspondingly. Thus, at each iteration of MLSGA, an initial charging path is constructed by a greedy method, and the novel local search operators are adopted to improve the quality of the initial charging path at the upper level. Charging time optimization is performed at the lower level by leveraging a Genetic Algorithm.   
 The second algorithm MTBCS based on a nested evolutionary strategy, where a hybrid of the Genetic Algorithm and the Local Search operators are adopted to speed up the charging path optimization at the upper level. Each feasible candidate of the upper level becomes an input for optimizing charging time at the lower level.
However, instead of optimizing charging time for all charging paths at the upper level, we divided the charging paths into groups and only chose a representative candidate to identify an optimal charging time at the lower level. Multiple charging times linked to the chosen charging paths will be optimized simultaneously by the Multitasking-CMA-ES algorithm to significantly reduce the running time compared to the traditional nested algorithm. 



The major contributions of this paper can be summarized as follows:
\begin{itemize}

\item Study the problem of minimizing energy depletion in \gls{wrsn} based on a novel partial charging approach that optimizes both the charging path and charging time simultaneously.
\item Provide network model, charging model, and a mathematical formulation of the investigated problem.
\item Propose two approximate algorithms to solve the investigated problem based on a bi-level optimization approach. The first algorithm adopts a multi-restart mechanism to explore multiple feasible points in the enormous search space. An initial charging path is constructed by a greedy method and local search operator to enhance the quality of the charging path. The second algorithm leverages a nested evolutionary strategy, where multiple charging times linked to the chosen charging paths at the upper level will be optimized simultaneously by the Multitasking and Covariance Matrix Adaptation Evolution Strategy.
\item Perform statistic analysis and extensive experiments to compare the efficiency of the proposed algorithms to the most related works on various network scenarios.

\end{itemize}

The rest of the paper is organized as follows. In \Cref{sec:related}, we highlight related works. \Cref{sec:model} presents the network model and problem formulation. Our proposed algorithms are described in \Cref{sec:propose_1} and \Cref{sec:propose_2}. Statistic analysis and experimental results on benchmark datasets are discussed in  \Cref{sec:evaluation}. Finally, conclusions and future works are provided in \Cref{sec:conclusion}.

\label{sec:intro}

\section{Literature review}
\label{sec:related}
Since the limited energy constraint has shown to be a bottleneck phenomenon in traditional \glspl{wsn}, many efforts have been devoted to solving the energy depletion avoidance problem. The literature can be classified into two primary methods to prolong the sensor lifetime, including energy-saving and energy replenishing \cite{tian2020charging}. The first method aims to minimize the sensor's energy consumption using various techniques such as deploying relay nodes, relying on cross-layer design, or using mobile sinks \cite{le2019node}. Although these methods may extend the sensor lifetime for a period, sensor energy depletion is unavoidable. In recent times, Wireless Rechargeable Sensor Networks (\glspl{wrsn}) that leverage the advantages of the wireless power transfer technology have emerged as a potential solution to the energy issue in traditional \glspl{wsn}. In \glspl{wrsn}, the sensor's lifetime mainly depends on the charging scheme of the Mobile Charger (\gls{mc}). Thus the most challenging problem in solving the \glspl{wrsn} energy issue is to find an effective charging scheme for the \gls{mc} to charge the sensor nodes. Most existing researches address this problem by focusing on two main approaches: charging path optimization and charging time optimization. 

Regarding the first approach, the authors in \cite{lyu2019periodic, shi2011renewable} studied the problem of maximizing the \gls{mc} vacation time at the depot (i.e., the time the \gls{mc} replenishes its battery). The authors proved that the optimal charging path of the MC is equivalent to the shortest Hamiltonian cycle. In the literature \cite{lyu2019periodic}, Lyu et al. proposed an enhanced periodic charging scheme of \cite{shi2011renewable} by taking into account the limited traveling energy of the MC and the imbalanced energy consumption among sensor nodes in the network. The authors addressed three possible network scenarios and applied a hybrid meta-heuristic algorithm based on Particle Swarm Optimization and Genetic Algorithm to determine the charging scheme for each scenario. Many researchers have worked on an on-demand charging strategy where the sensors send a charging request whenever their energy level drops below a threshold value. The MC charging decision will be made based on these requests. In \cite{peng2010prolonging} the authors constructed a proof-of-concept prototype of the system and attempted to solve the charging path planning problem by maximizing the network lifetime using a greedy algorithm depending on the lifetime of each sensor node. The studies of \cite{feng2016starvation,zhu2018adaptive} further complete the work in \cite{peng2010prolonging} by proposing a Starvation Avoidance Mobile Energy Replenishment scheme (SAMER) that adapt well to the high diversity of energy consumption but also fully consider the fairness of charging response \cite{feng2016starvation} and an Invalid Node Minimized Algorithm (INMA) to optimize the waiting queue \cite{zhu2018adaptive}. 
Aiming to find the optimal charging plan for the MC, Lin et al. in \cite{lin2018mpf} studied the issues of both the fully charging strategy (the MC charges all the sensors to maximize their battery) and the partial charge plan (MC only charges a part of the sensor battery capacity). They proposed a Mixed Partial and Full charge plan (named MPF), including three specialized modules, i.e., the evaluation module, adjustment module, and selection module, to overcome the disadvantages of both charging plans. 
In order to enhance the charging utility, the authors in \cite{fu2015optimal, yang2019igrc} used a multi-nodes charging model that allows multiple sensors can be charged at the same time. Besides, They also studied how to find a collection of charging locations to maximize the MC charging power. These charging locations are determined by dividing the network charging field into smaller parts using a variety of techniques such as Smallest Enclosing Space \cite{fu2015optimal} or the grid \cite{ yang2019igrc}.

Some researchers have been working on the optimization of the charging time. Regarding this problem, Xu et al. in \cite{xu2016maximizing} proposed a novel idea to divide the charging energy and time into unit slots, and then the problem can be reduced to a matching problem between sensors and time slots. They then solve this problem by using the Maximum weighted matching in a bipartite graph to find the optimal charging schedule. Whereas the authors in \cite{nguyen2020extending} assumed that a charging path is predetermined and proposed a lightweight greedy algorithm to determine the optimal charging time at each charging location. Their ultimate goal is to maximize the network lifetime. The term ``\textit{network lifetime}" is defined by the interval from when the network starts until the first sensor node is dead (due to the exhausted energy). The authors in \cite{carrabs2020optimization} handled the problem of deciding the amount of the charged energy for each sensor node to maximize the time of target monitoring.\\

\indent An effective charging scheme should be considered both the charging path and charging time of the \gls{mc}. The existing works, however, only deal with these factors separately. 
The authors of \cite{huong2020genetic,huong2020optimizing} optimize both factors of the charging scheme by decomposing the problem into two sub-problems: finding the optimal charging path and determining the optimal charging time at each sensor. Then, each subproblem can be solved using an approximate approach based on the genetic algorithm \cite{huong2020genetic}, or a mixed-integer linear programming model \cite{huong2020optimizing}. This separation of the two phases could lead to the result of the first phase not being optimized for the final solution.  In this paper, we propose the bi-level optimization approach to simultaneously optimize both the charging path and the charging time to minimize the number of the dead sensor nodes in the \glspl{wrsn}.
\\
\indent 
\indent In the past decade, Covariance Matrix Adaptation Evolutionary Strategy (CMA-ES) has been shown to be a state-of-the-art algorithm for continuous optimization problems with fast convergence speed. As CMA-ES belongs to Evolution Strategy (ES), the algorithm consists of three essential operators: recombination, mutation, and selection. Furthermore, it also possesses appealing characteristics such as derivative-free, covariant, off-the-shelf, scalable. It is advantageous on problems that are non-convex, non-separable, ill-conditioned, multi-modal, and noisy evaluations \cite{auger2012tutorial}. Because of these advantages, we decided to adopt the CMA-ES for the charging time optimization for one of our proposed algorithms, and details could be found in section \ref{algo:lower-level} \\

\section{Network system setting and problem formulation}
\label{sec:model}

		\begin{table}[!ht]
	    \caption{List of main notations}
	    \label{tab:notation}
	    \begin{tabular}{|l|m{0.76\textwidth}|}
	       \hline
	        \textbf{Notation}   & \textbf{Description} \\
	        \hline
	      $V= \{1, ..., n\}$ & a set of static sensor nodes \\
	       $ 0 $  & the base station \\
	       $ e^{max}, e^{min} $ & energy capacity and minimum energy level of sensor battery \\
	       $ p_{i} $ & average energy consumption rate of sensor $i$ \\
	        $ e^{init}_{i}, e^{depot}_{i} $ & initial energy and residual energy of node $i$ at the beginning and at the finishing of the scheduling period, respectively
	        \\
	       $ e_{i} $ & residual energy of sensor $i$ when \gls{mc} arrives \\
	        $ d_{i,j} $  &  euclidean distance  between two sensors $i$ and $j$ \\
	        $a_{i}$ & the time that the mobile charger arrives at sensor node $i$\\
	          $ z_{i} $ & a binary variable represents the active status of sensor $i$ \\
    	       $ \Delta_{i} $ & energy reduction of sensor $i$ in a charging cycle (e.g. the energy level gain of $i$ between two timings: beginning and finishing time of the scheduling period T) \\
    	        $ E_{\gls{mc}} $ & maximum energy level of the \gls{mc} in a charging cycle \\
    	        
	       $E_{move}, E_{charge}$ & total consumed energy of \gls{mc} for traveling and charging processes.  \\
    	    $ P_M, U$ & the per-second energy consumption rate of \gls{mc} when traveling and when charging\\
    	    $v$ & the \gls{mc}’s velocity \\
	        $T$ & the scheduling period of a charging cycle\\
	        $T_{travel}, T_{charge}$ & the total time for traveling and charging, respectively.  \\
	       $\boldsymbol{\wp}$ & the discrete vector representing a charging path of \gls{mc}\\
	        $\mathcal{T}$ & the continuous vector representing a time charging sequence at all charging locations\\
    	   
    	       \hline
	    \end{tabular}
	\end{table}
	\subsection{Network Setting}
We consider a Wireless Rechargeable Sensor Network deployed over a region of interest. The network model can be represented as a weighted graph $\mathbf{G} = (\mathbf{V} 
, \mathbf{E}, \mathbf{D})$, where $\mathbf{V} = \{0, 1, 2, ..., n\}$ includes a set of static sensors (nodes 1, 2, ..., n) that are equipped with a wireless receiver as well as a base station (node 0) responsible for gathering data from the sensor nodes. $\mathbf{E}$ represents a set of edges between any two nodes that are in the range of their communication. The set $\mathbf{D} \subseteq V \times V $ represents the travel distances following Euclidean formulation between sensor nodes. Each sensor consumes energy for three essential tasks including sensing, receiving, and transmitting, determined by the network topology. In this system, we leverage the energy consumption model presented in \cite{heinzelman2002application}.
\begin{itemize}
    \item The energy consumption of a sensor for receiving $l$ bits data with distance $d$ is calculated as follows: 
    \begin{ceqn}
         \begin{equation}
             \label{eq:e_r}
            er ~= ~l\epsilon_{elec}     ~\text{(J/b)},
        \end{equation}
         \end{ceqn}
     \item The energy consumption of a sensor for transmitting $l$ bits data with distance $d$ is defined as follows: 
      \begin{ceqn}
    \begin{equation}
    	\label{eq:e_t}
    	et = 
    	\begin{cases}
    		l\epsilon_{elec} + l \epsilon_{fc}\times d^{2} ~\text{if} ~ d < d_0 \\
    		l\epsilon_{elec} + l \epsilon_{mp}\times d^{4} ~\text{if}~ d \geq d_0,
    	\end{cases} 
    \end{equation}
    \end{ceqn}
\end{itemize}

where the $\epsilon_{elec}$ is the electronics energy expenditures per bit data to run the transmitter or receiver. $\epsilon_{fc}$ and $\epsilon_{mc}$ are the amplifier energy of transmitting one bit of data in free space ($d^2$) and multi-path fading ($d^4$) model, respectively. $d_0$ represents the threshold distance between the receiver and transmitter, respectively. The parameters in the network model are set as: $\epsilon_{elec} = 0.05J/bit$, $\epsilon_{fc} = 0.01J/bit/m^2$, $\epsilon_{mp}=0.013pJ/bit/m^4$, $d_0 = \sqrt{\dfrac{\epsilon_{fc}}{\epsilon_{mp}}}$.    

\begin{figure}[ht]
    \centering
    \includegraphics[width= 0.5\textwidth]{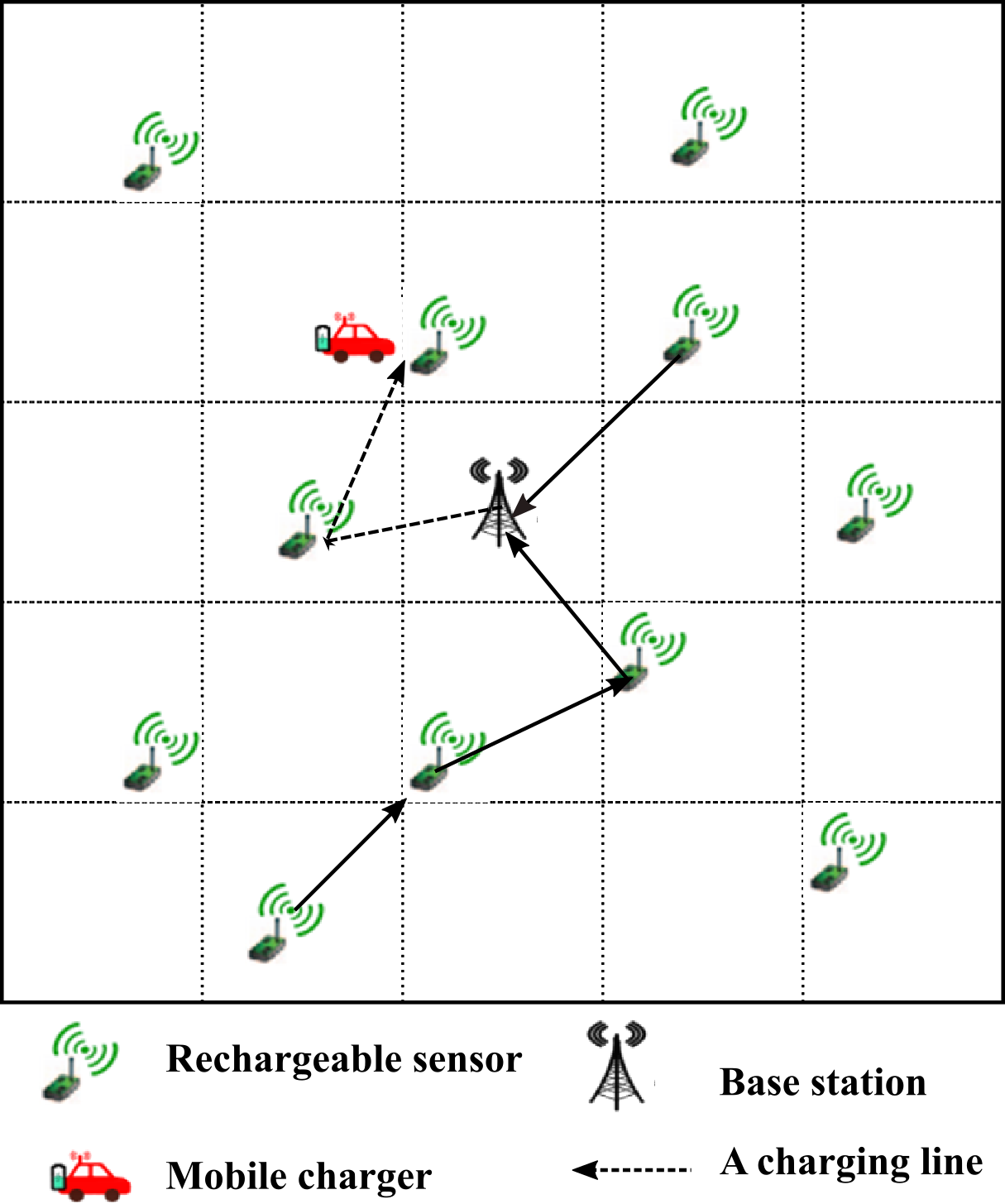}
    \caption{A wireless rechargeable sensor network system.}
    \label{fig:wrsn}
\end{figure}

The sensors periodically send the data packets containing the sensory data, their residual energy, and the time stamp to the \gls{bs} through the multi-hop transfer protocol. From the obtained information, the \gls{bs} can estimate the average energy consumption $e_i$ and the remaining energy ($e^{init}_i$) of sensor $i$ by using methods in \cite{zhu2018adaptive}. Each sensor has the battery capacity $e^{max}$ and minimum operating energy threshold $e^{min}$. When the energy of sensor is less than $e^{min}$, it become a dead sensor and can not perform its tasks. Therefore, to avoid the energy depletion of sensors, a Mobile Charger (MC) is adopted. In such context, the MC departs at the \gls{bs} (also plays a role as the depot) and periodically travels to the location of sensor nodes in the network to replenish their battery, in which for each sensor node $i$, the \gls{mc} will spend an interval $t_i$ to charge it.
After finishing the charging process for all sensor nodes, the MC returns to the \gls{bs} to fully recharge its battery and prepare for the next charging round. The network system is illustrated in Figure \ref{fig:wrsn}.

	\subsection{The problem formulation}
	\label{subsect:formulation}
Without loss of generality, a charging cycle period is set to $T$, and the MC replenished itself to its maximum battery capacity $E_{MC}$ for beginning the charging round. 
In this work, we study how to determine the optimal charging path and time to minimize energy depletion of sensors in \glspl{wrsn}. We name the investigated problem as \gls{edmp}. Formally, a feasible solution to the \gls{edmp} consists of two factors: a charging path $\wp = (\pi_0, \pi_1, \pi_2, ..., \pi_n, \pi_{n+1})$ and a charging time sequence $\mathcal{T} = (\tau_{\pi_1}, \tau_{\pi_2}, \dots, \tau_{\pi_n})$; where $\{\pi_1, \pi_2, ..., \pi_n\}$ is a permutation of the set $\{1, 2, ..., n\}$ and $\pi_0 \equiv \pi_{n+1} \equiv 0$. $\tau_{\pi_i}$ indicates the charging time at sensor $\pi_i$ when the \gls{mc} travels following the path $\wp$. To easy in the solution representation, we denote a charging path $\pi_o \rightarrow  \pi_1 \rightarrow \pi_2 \rightarrow \pi_3 ... \rightarrow \pi_n \rightarrow \pi_{n+1}$  as a vector ($\pi_1, \pi_2,\pi_3,\pi_n$).

Since the MC always expenses energy during the functioning process, MC's total energy for traveling and charging processes must not exceed its maximum battery capacity. We have: 
\begin{equation}
    \label{const_EMC}
    T_{travel} \times P_M + \sum_{i = 1}^{n} \tau_{\pi_i} \times U \leq E_{MC} 
    \end{equation}
where, $P_M$ and $U$ represent the per-second energy consumption rates of MC when traveling and charging, respectively. $T_{travel} = \dfrac{\sum_{j=0}^{n}d_{\pi_{j}\pi_{j+1}}}{v}$ is the total traveling time of MC with velocity v.

In addition, the total traveling time and charging time does not also exceed the scheduling period $T$:
\begin{equation}
    \label{const_T}
    T_{travel} ~ + ~ T_{charge} \leq T
\end{equation}
where, $T_{charge} = \sum_{i = 1}^{n} \tau_{\pi_i} $ is the total charging time that the MC spends to charge all sensors in one charging cycle.

\begin{figure}[h]
	\centering
	\includegraphics[width=0.7\textwidth]{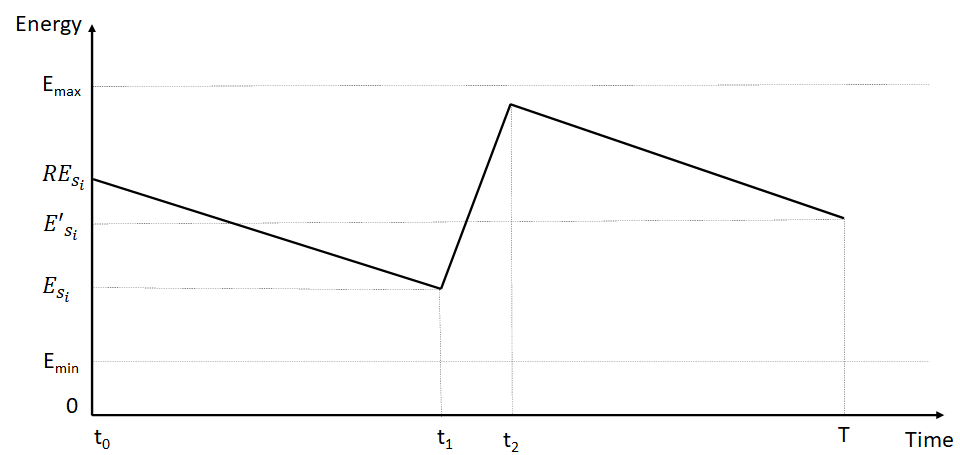}
	\caption{ Energy fluctuation of a sensor in a charging cycle $T$, where $t_0$ is the timing that $MC$ starts at the depot. Energy of the sensor decreases in the first phase ($t_0$ to $t_1$) and the third phase ($t_2$ to $T$). The second phase ($t_1$ to $t_2$) is the timing that the sensor's battery is replenished.}
	\label{fig:energy_flutuationOfsensor}
\end{figure}

During the cycle period $T$,  we have observed that the energy fluctuation of $\pi_i$ may be divided into three major phases: the pre-charging phase, the in-charging, and the after-charging phase. For the first phase, since \gls{mc} travels and charges the sensors $\{\pi_1, \pi_2, \ldots, \pi_{i-1}\}$, $\pi_i$ has to wait a period ò time. As a result, the energy of $\pi_i$ decreases gradually. For the second phase, the energy of $\pi_i$ is increased by receiving energy from the \gls{mc}. Finally, in the third phase, when \gls{mc} leaves $\pi_i$ and travels for the other sensors, the energy of $\pi_i$ again declines for performing its tasks. Fig. \ref{fig:energy_flutuationOfsensor} illustrates the energy fluctuation of sensor $\pi_i$ during the charging cycle.

Let $e_{\pi_i}$ and $e^{depot}_{\pi_i}$ are the residual energy of sensor $\pi_i$ at two timings: the time that \gls{mc} arrives at sensor $\pi_i$ and the the end of period $T$, respectively.  
Accordingly, $e_{\pi_i}$ and $e^{depot}_{\pi_i}$ are calculated as follows:         
\begin{equation}
    \label{constrant_e_i}
   e_{\pi_i} = e^{init}_{\pi_i} - \Bigg(
   \sum_{j=0}^{i-1}\dfrac{d_{\pi_j\pi_{j+1}}}{v} + \sum_{j=0}^{i-1}t_{\pi_j}\Bigg) \times p_{\pi_i}
\end{equation}
and 
\begin{equation}
    \label{constrant_e_depot_i}
   e^{depot}_{\pi_i} = e^{init}_{\pi_i} - T \times p_{\pi_i}
\end{equation}
The first term $e^{init}_{\pi_i}$ in formula (\ref{constrant_e_i}) and (\ref{constrant_e_depot_i}) is the initial energy of sensor $\pi_i$ at each charging cycle, while the second term in both formulas above indicates the energy that the sensor $\pi_i$ expenses from when MC starts the charging cycle until it arrives at the location of $\pi_i$ and when finishing the charging cycle at \gls{bs}, respectively.  
Let $z_{\pi_i}$ be a binary variable which represents the status of sensor $\pi_i$, where $z_{\pi_i}$ = 1 if $\pi_i$ dies in charging cycle, otherwise $z_{\pi_i}$ = 0.   
In Fig. \ref{fig:energy_flutuationOfsensor}, the remaining energy of sensor $\pi_i$ attains the lowest value in a charging cycle at two timings, either when \gls{mc} arrives at $\pi_i$ or the end of the charging cycle. Thus, to determine whether sensor $\pi_i$ dies in the charging cycle or not, we only consider the remaining energy of $\pi_i$ at the above two timings. It means that:
 \begin{ceqn}
		\label{eq:status}
		\begin{equation}
			z_{\pi_i} = 
			\begin{cases}
				1~\text{if}~e_{\pi_i} < e^{min}~ \text{or} ~e^{depot}_{\pi_i} < e^{min} \\
				0~\text{otherwise},
			\end{cases}
		\end{equation}
	\end{ceqn}

Intuitively, to avoid energy depletion, sensors should be recharged before they drain energy. Thus, we need to determine the optimal charging path sequence through all sensors and the corresponding charging time that MC spends at each sensor to minimize the number of sensor nodes depleted their energy (named dead sensor nodes). Moreover, our objective focuses on minimizing the number of dead sensor nodes at the current charging round and the next charging round. Thus, the problem objective is defined as follows:
\begin{ceqn}
    \begin{equation}
			\label{objFunc} f = \alpha \frac{\sum_{i=1}^n z_{\pi_i}}{n}+ (1-\alpha) \frac{max_{\pi_i \in V }\{\Delta_{\pi_i}\}}{e^{max} - e^{min}} \rightarrow min,
	\end{equation}
\end{ceqn}
where, $\alpha$ is a control parameter and $\Delta_{\pi_i}$ is the energy difference of the sensor $\pi_i$ between two timings when \gls{mc} starts and finishes the charging cycle. The value of $\Delta_{\pi_i}$ is defined by:
		\begin{ceqn}
		    \begin{equation}
	\label{eq:energy_reduction_define}
			\Delta_{\pi_i}=
			\begin{cases}
				e^{init}_{\pi_i} - e^{depot}_{\pi_i}, & \text{if $z_{\pi_i} = 0$ and $e^{init}_{\pi_i} 	> 	e^{depot}_{\pi_i}$} \\
				0, & \text{otherwise.} \\
			\end{cases}
	\end{equation}
		\end{ceqn}
	
The first component in the objective function \ref{objFunc} depicts the ratio of the dead sensor (due to energy depletion) over the total network number of sensors and favors the minimum when the number of the dead sensors at the current charging cycle is the smallest. While the second component tends to maximize the sensor's remaining energy after each charging cycle toward minimizing the dead nodes in the next charging cycle. Table \ref{tab:notation} describes the notations used in this paper.

The optimization problem \gls{edmp} can be represented as follows.

\begin{ceqn}
    \begin{equation}
    \label{object_function}
        Min_{\boldsymbol{\wp}, \boldsymbol{t}} ~  \Bigg(\alpha\frac{\sum_{i=1}^n z_{\pi_i}}{n}+ (1-\alpha)\frac{max_{\pi_i \in V }\{\Delta_{\pi_i}\}}{e^{max} - e^{min}} \Bigg )
    \end{equation}
\end{ceqn}
 \indent\textit{ s.t. Constraints (\ref{const_EMC}), (\ref{const_T}),}
    \begin{align}
		\label{const_tau_max}
            &e^{init}_{\pi_i} + \tau_{\pi_i} \times (U -p_{\pi_i}) \leq e^{max}  ~~~~~\forall \pi_i \in V \\
		 \label{const_t_positive}
		  &\tau_{\pi_i} \ge 0 ~~~~~\forall \pi_i \in V 
    \end{align}

Constraint (\ref{const_tau_max}) indicates that the total charged energy of each sensor does not exceed the sensor's maximum battery capacity. Constraint (\ref{const_t_positive}) expresses the charging time at each sensor node must be not negative.

The \gls{edmp} is considered as a Bi-level Optimization Problem (BLOP) because \gls{edmp} consists of two optimization levels, including the charging path level and the charging time level, referred to as upper-level and lower-level optimization tasks, respectively. Moreover, the charging time optimization task is nested as a constraint of the charging path optimization task. Thus, \gls{edmp} is known as the NP-hard problem \cite{sinha2017review}. It is difficult to find an exact solution in large-scale scenarios. Therefore, we adopt approximate methods based on bi-level optimization approach to deal with the investigated problem.   

We have an essential observation that the total charging time of the MC significantly affects the performance of a charging scheme. If the \gls{mc} spend too much time for the charging process, the period of one charging cycle may substantially increase lead to the energy depletion of critical sensors before they are recharged in the next charging round. In addition, the energy of MC maybe not be enough for the \gls{mc} to go back to the depot. However, too little charging time by the MC may lead to the sensors energy level too low and not enough to survive until being charge in the following cycle. So, before representing the proposed algorithm, in next section, we will determine the upper bound of the total charging time in one charging round.  

\subsection{Total charging time determination}
	
	We first construct the upper bound of the total charging time as follows. Suppose that a scheme with a charging tour $\wp$ =  ${\pi_0\rightarrow\pi_1\rightarrow\pi_2\ldots\rightarrow\pi_n\rightarrow\pi_{n+1}}$ is performed by MC in the charging round $k^{th}$. 
	Assuming that all the sensors are still active after the charging round. That assumption can be formulated as:
	\begin{equation}
		\label{con:ub1}
		\sum_{i=1}^{n}e^{init}_{\pi_i}+\sum_{i=1}^{n}\tau_{\pi_i}\times U - n\times e^{min} \geq (T_{charge} + T_{travel}) \times \sum_{i=1}^{n}p_{\pi_i}
	\end{equation}
	That means the sum of the initial energy of the network and the extra energy that the network receives from MC must be greater or equal to the total energy consumption by all the sensors. Otherwise, the sensors battery can be drained at the end of the charging round.	Replace $T_{charge}$ into the upper equation, the constraint (\ref{con:ub1}) can be rewritten as follows:
	\begin{equation}
		\label{con:ub2}
		\sum_{i=1}^{n}e^{init}_{\pi_i}-n\times e^{min}-T_{travel} \times \sum_{i=1}^{n}p_{\pi_i} \geq 	T_{charge}\times \left( \sum_{i=1}^{n}p_{\pi_i} - U\right)
	\end{equation}
	From (\ref{con:ub2}), if $U < \sum_{i=1}^{n}p_{\pi_i}$  we have:
	\begin{equation}
		\label{ub:chargingtime1}
		T_{charge} \leq \frac{\sum_{i=1}^{n}e^{init}_{\pi_i}-n\times e^{min} - T_{travel} \times \sum_{i=1}^{n}p_{\pi_i}}{\sum_{i=1}^{n}p_{\pi_i} - U} 
	\end{equation}
	Besides, because the energy of a sensor can not exceed its battery capacity, so we have
	\begin{equation}
		\label{con:ub3}
		\sum_{i=1}^{n}e^{init}_{\pi_i} + \sum_{i=1}^{n}\tau_{\pi_i}\times U-(T_{charge} + T_{travel})\times\sum_{i=1}^{n}p_{\pi_i}\le n\times e^{max}
	\end{equation}	
	Replace $\sum_{i=1}^{n}\tau_{\pi_i}$ by $T_{charge}$ and rewrite the constraint (\ref{con:ub3}), we have
	\begin{equation}
		\label{con:ub4}
		T_{charge}\times (U-\sum_{i=1}^{n}p_{\pi_i})\le n\times e^{max}-	\sum_{i=1}^{n}e^{init}_{\pi_i} + T_{travel} \times\sum_{i=1}^{n}p_{\pi_i}
	\end{equation}	
	From constraint (\ref{con:ub4}), if $U > \sum_{i=1}^{n}p_{\pi_i}$ we have:
	\begin{equation}
		\label{ub:chargingtime2}
		T_{charge} \leq \frac{n\times e^{max}-	\sum_{i=1}^{n}e^{init}_{\pi_i} + T_{travel} \times\sum_{i=1}^{n}p_{\pi_i}}{U-\sum_{i=1}^{n}p_{\pi_i}}, \ 
	\end{equation}
	In the case $U = \sum_{i=1}^{n}p_{\pi_i}$, i.e. the energy consumed by the whole network is totally compensated by the charging mobile MC, the equations (\ref{ub:chargingtime1}) and (\ref{ub:chargingtime2}) are obviously satisfied. So, the upper bound of the charging time just needs to follow the constraint (\ref{const_EMC}), which is rewritten as
	\begin{equation}
		\label{ub:chargingtime3}
		T_{charge} \leq 	\frac{E_{MC} - T_{travel} \times P_M}{U}
	\end{equation}
	From (\ref{const_EMC}), (\ref{ub:chargingtime1}), (\ref{ub:chargingtime2}), and (\ref{ub:chargingtime3}), we can calculate the upper bound of total charging time for the charging plan as follows:
	\begin{equation}
		\label{ub:chargingtime}
		T_{charge} \le 
		\begin{cases}
			min\left(\frac{E_{MC} - T_{travel} \times P_M}{U}, \frac{\sum_{i=1}^{n}e^{init}_{\pi_i} - n\times e^{min} - T_{travel} \times \sum_{i=1}^{n}p_{\pi_i}}{\sum_{i=1}^{n}p_{\pi_i} - U}\right), & \text{if $U < \sum_{i=1}^{n}p_{\pi_i}$}\\
			min\left(\frac{E_{MC} - T_{travel} \times P_M}{U}, \frac{n\times e^{max}-	\sum_{i=1}^{n}e^{init}_{\pi_i} + T_{travel} \times\sum_{i=1}^{n}p_{\pi_i}}{U-\sum_{i=1}^{n}p_{\pi_i}}\right), & \text{if $U > \sum_{i=1}^{n}p_{\pi_i}$} \\
			\frac{E_{MC} - T_{travel} \times P_M}{U}, & \text{if $U = \sum_{i=1}^{n}p_{\pi_i}$}
		\end{cases}
	\end{equation}

	It is clear that to prolong sensors' lifetime; they should be charged as much as possible. In other words, the energy that MC uses to charge for the network should be as much as possible. Based on the above observation and the equation (\ref{ub:chargingtime}), in this paper, we determine the total charging time for a given charging tour ${\pi_0\rightarrow\pi_1\rightarrow\pi_2\ldots\rightarrow\pi_n\rightarrow\pi_{n+1}}$ as follows.
	\begin{equation}
		\label{sumTc}
		T_{charge} = 
		\begin{cases}
			min\left\{\frac{E_{MC} - T_{travel} \times P_M}{U}, \frac{\sum_{i=1}^{n}e^{init}_{\pi_i}-n\times e^{min}-T_{travel} \times \sum_{i=1}^{n}p_{\pi_i}}{\sum_{i=1}^{n}p_{\pi_i} -  ~U}\right\}, & \text{if $U < \sum_{i=1}^{n}p_{\pi_i}$}\\
			min\left\{\frac{E_{MC} - T_{travel} \times P_M}{U}, \frac{n\times e^{max}-	\sum_{i=1}^{n}e^{init}_{\pi_i} + T_{travel} \times\sum_{i=1}^{n}p_{\pi_i}}{U-\sum_{i=1}^{n}p_{\pi_i}}\right\}, & \text{if $U > \sum_{i=1}^{n}p_{\pi_i}$} \\
			\frac{E_{MC} - T_{travel} \times P_M}{U}, & \text{if $U = \sum_{i=1}^{n}p_{\pi_i}$}
		\end{cases}
	\end{equation}

\section{Hybrid of multi-start local search and genetic algorithm}
\label{sec:propose_1}

\subsection{Motivation of proposed algorithm}
The work in \cite{huong2020genetic} proposes a two-phase algorithm in which the former phase tries to find an optimal charging path and the latter phase focuses on optimizing the charging time while assuming that MC follows the path obtained by the first phase. However, the result of the first phase may not be an optimal path, and thus the second phase may be misled into an incorrect assumption. From that observation, in this section, we propose an algorithm for constructing the charging schedule, in which several (local optimal) paths are considered to optimize the charging time.

Due to the ability to produce multiple locally optimal solutions by running a local search procedure from multiple starting points, the multi-start local search algorithm is chosen to optimize the charging path. In addition, the genetic algorithm, with exemplary performance in global search, is adopted for determining the charging time at each sensor. A novel local search operator is also proposed in this section to improve the quality of charging paths. Instead of determining the charging time for only one path as in \cite{huong2020genetic}, in each iteration of the proposed algorithm, we attempt to find an optimal path, and then the corresponding charging time for that path is determined.  
To simplify the presentation, the proposed algorithm in this section is named Hybrid of Multi-start Local Search and Genetic Algorithm (MLSGA). 

\subsection{ MLSGA algorithm scheme}
The outlines of our approach are presented in Algorithm \ref{algo:hmls-ga}. Each iteration of the algorithm consists of two steps. The first step aims to construct an optimal path for MC. The charging time for the path obtained after the first step is then determined in the second step by using a genetic algorithm (line 7). After two steps, the new charging schedule $(\wp, \mathcal{T})$ will be compared with the best overall solution obtained so far to see if a new best is found. If this is the case, we successfully replace the best overall solution with the improved solution (lines 8-10). In the end, the best overall solution is returned as the final result.

\begin{algorithm}[H]
	\label{algo:hmls-ga}
	\caption {\textbf {Hybrid Multi-start Local Search and Genetic Algorithm}}
	\KwIn{A wireless rechargeable sensor network consists of $n$ sensors $\{1, 2,\ldots, n\}$ and a mobile charger.}
	\KwOut{ A charging path $\wp^* = (\pi^*_1,\pi^*_2\ldots\pi^*_n\ )$  and \\ ~~~~~~~ ~~~~ ~~a charging time vector $\mathcal{T}^* = (\tau_{\pi^*_1},\tau_{\pi^*_2},\ldots,\tau_{\pi^*_n})$ correspondingly.}
	\Begin{
		$\wp^* \longleftarrow null$\;
		$\mathcal{T}^* \longleftarrow null$\;
		\While{terminate conditions are not satisfied}{
			$\wp_0 \longleftarrow$ construct an initial charging path\tcp*{see Subsect. \ref{algo:mlsga:step1}}
			$\wp \longleftarrow$ local search on $\wp_0$ until a local optima is reached \tcp*{using evaluation method in \ref{algo:greedy_eval_method} and local search operator in \ref{algo:local_search_operator}}
			Determine the optimal charging time vector $\mathcal{T}$ corresponding to the charging path $\wp$ using Genetic Algorithm \tcp*{see in Subsect. \ref{algo:mlsga:step2}}
			\If{$f(\wp^*, \mathcal{T}^*) > f(\wp, \mathcal{T})$}{
				$\wp^* \longleftarrow \wp$\;
				$\mathcal{T}^* \longleftarrow \mathcal{T}$\;
			}
		}
		\textbf{return} $\{\wp^*, \mathcal{T}^*\}$\;
	}
\end{algorithm}

Before presenting two steps of each iteration, we first introduce a greedy method for charging path evaluation and a local search operator for improving a given path.

\subsection{Charging path evaluation method}
\label{algo:greedy_eval_method}
To precisely evaluate a given charging path, the corresponding optimal charging time must be determined. Unfortunately, finding the optimal charging time is not straightforward itself. Thus, to evaluate all generated paths during the search process, we propose an evaluation method by greedy determining the charging time.

For a given charging path $\wp \equiv \pi_0\rightarrow\pi_1\rightarrow\pi_2\ldots\rightarrow\pi_n\rightarrow\pi_{0}$, intuitively, the charging time at a sensor depends on its energy status and energy consumption rate, which means sensors with less residual energy and higher consumption rate should be charged more than the others. Based on that observation, the charging time is greedily determined as follows. First, we define for each sensor $\pi_i$ a weight factor as 
	\begin{equation}
		w_{\pi_i}=max\left (0,~ \frac{p_{\pi_i}}{\sum_{i=1}^{n}p_{\pi_i}}- 	\frac{e^{init}_{\pi_i}}{\sum_{i=1}^{n}e^{init}_{\pi_i}}\right ).
	\end{equation}
	After determining the weights of all sensors, the charging time at sensor $\pi_i$ is assigned as follows
	\begin{equation}
		\label{tau_greedy}
		\tau_{\pi_i} = T_{charge} \times \frac{w_{\pi_i}}{\sum_{i=1}^{n}w_{\pi_i}} \, ,
	\end{equation}
where $T_{charge}$ is the total of charging time of MC in the charging round that is calculated by \ref{sumTc}. The charging path $\wp$ is then evaluated with the above charging time under the objective function \eqref{object_function}.

\subsection{Local search operator for charging path improvement}
\label{algo:local_search_operator}
In this subsection, we propose local search operators to improve the quality of a charging path. Given a schedule that consists of a path $\wp = (\pi_1,\pi_2\ldots,\pi_n)$ and charging time at each sensor $\mathcal{T} = (\tau_{\pi_1}, \tau_{\pi_2}, \ldots, \tau_{\pi_n})$, following \ref{eq:status}, there are two reasons that cause the dead of a sensor $\pi_i$ as follows:
\begin{itemize}
    \item $\pi_i$ is dead before being charged ($x_{\pi_i} = 1$ because $e_{\pi_i} < e_{min}$).
    \item $\pi_i$ is dead after being charged, i.e., the total initial energy of $\pi_i$ and the energy received from MC is not enough to keep it active to the end of the charging round ($y_{\pi_i} = 1$ cause of $e^{depot}_{\pi_i} < e_{min}$).
\end{itemize}

Let define $D_B$ and $D_A$ as two collections of any dead sensor caused by the first and second reasons. The basic idea of our local search operator is as follows.

If a sensor $\pi_i \in D_B$,  $\pi_i$ is charged too late. So, we move it forward to the front of the charging sequence. By doing that, the charged waiting time of $\pi_i$ will be reduced. Conversely, if a sensor $\pi_i \in D_A$, the charging time that the MC spends at the location of $\pi_i$ should be increased. From Equations \eqref{constrant_e_i} and \eqref{const_tau_max}, the charging time at sensor $\pi_i$ has an upper bound value as follows:
	\begin{equation}
		\label{tau_max2}
		\tau^{max}_{\pi_i} = \frac{e^{max} - 	e^{init}_{\pi_i}+\Bigg(\sum_{j=0}^{i-1}\dfrac{d_{\pi_j,\pi_{j+1}}}{v}+\sum_{j=1}^{i-1}\tau_{\pi_j}\Bigg)\times{p_{\pi_i}}}{U - p_{\pi_i}}.
	\end{equation}
	From \eqref{tau_max2}, to increase $\tau^{max}_{\pi_i}$, we need to increase the term $\sum_{j=0}^{i-1}\dfrac{d_{\pi_j,\pi_{j+1}}}{v}+\sum_{j=1}^{i-1}\tau_{\pi_j}$, i.e., the sensor $\pi_i$ should be charged later. Thus, we move $\pi_i$ to a backward position of the charging queue $\wp$. Algorithm \ref{algo:ls_operator} represents the pseudo-code of the proposed local search operator.

	\begin{algorithm}[H]
		\label{algo:ls_operator}
		\caption {\textbf { Local search operator}}
		\KwIn{A charging path $P = (\pi_1, \pi_2\ldots,\pi_n)$ and charging time at each sensor $\mathcal{T} = (\tau_{\pi_1}, \tau_{\pi_2}, \ldots, \tau_{\pi_n})$}
		\KwOut{A neighbour charging path $\wp'$ of the charging path $\wp$.}
		
		\Begin{
			$NS1 \longleftarrow |D_B|$\; 
			$NS2 \longleftarrow |D_A|$\; 
			\uIf{NS1 + NS2 = 0}{
				$P' \longleftarrow$  random move a sensor in $\wp$\;
			}
			\uElseIf{$D_A$ is empty or $rand(0, 1) \le \frac{NS1}{NS1+NS2}$}{
				Choose a random sensor $\pi_i \in D_B$\;
			    $j \longleftarrow argmax_{j \in [1, i-1]} \{a_{\pi_j}: a_{\pi_j} \le l_{\pi_i}\}$\;
				$P' \longleftarrow$ move $\pi_i$ to a random position in range $[1, j]$\;
				
			}	
			\Else{
				Choose a random sensor $\pi_i \in D_A$\;
				$j \longleftarrow argmax_{j \in [i+1, n]} \{a_{\pi_j} : a_{\pi_j} \le l_{\pi_i}\}$\;
				$P' \longleftarrow$ move $\pi_i$ to a random position in range $[i+1, j]$\;
			
			}
			\textbf{return} $\wp'$\;
		}
		
	\end{algorithm}
	where $l_{\pi_i}$ is the lifetime of sensor $\pi_i$ without the replenishing energy,
	\begin{equation}
		l_{\pi_i} = \frac{e^{init}_{\pi_i} - e_{min}}{p_{\pi_i}},
	\end{equation}
	and $a_{\pi_j}$ is the time that MC arrives at $\pi_j$ and starts charging,
	\begin{equation}
		a_{\pi_j} = \sum_{j=0}^{i-1}\dfrac{d_{\pi_j,\pi_{j+1}}}{v}+\sum_{j=1}^{i-1}\tau_{\pi_j}.
	\end{equation}
In the above algorithm, if a sensor $\pi_i$ is dead, we move it to a new position based on its lifetime $l_{\pi_i}$ (lines 9, 13) to ensure that it will not die after the move. Two well-known operators, namely, Two-exchange and Relocated, are adopted with the same probability to move $\pi_i$ to another position. 
\subsection{Charging path construction}
\label{algo:mlsga:step1}
The charging path of the \gls{mc} is designed based on the observation: a charging schedule needs to have a low traveling cost to maximize the charging energy for sensor nodes. Considering the energy aspect, since the the MC energy is limited, minimizing the traveling energy to increase the received energy by the sensor nodes. Regarding the time aspect, the more optimized the charging tour be, the less time the sensors have to wait. Therefore, a greedy approach is used at each iteration of the proposed algorithm to generate the charging tours with a low moving cost as a good starting point.
		
Specifically, we use the well-known k-nearest neighbor algorithm \cite{kizilatecs2013nearest} to generate an initial charging path. The method starts with a partial solution that consists of the base station $\pi_0$, and a list of  $L$ is composed of $k$ unvisited nearest sensors evaluated by moving distance. Each subsequent step chooses a random sensor from the list $L$ to add to the partial solution. This technique generates these different charging paths for each iteration. For the diversity of the starting points, we randomly select the value of $k$ in the range [2, $n$] for each charging path, where $n$ is the number of sensors. If $k$ is small, the method generates an initial charging path with a low traveling cost. In contrast, If $k$ is close to $n$, the method generates an individual with a higher degree of randomness.

The local search operator proposed in Subsect. \ref{algo:local_search_operator} is then performed on the initial path until the local optima are reached.
\subsection{Charging time optimization based on genetic algorithm}
\label{algo:mlsga:step2}
Suppose that $\wp \equiv (\pi_1, \pi_2\ldots, \pi_n)$ is the charging path obtained after the first step of an iteration. This subsection represents how we determine the charging time at each sensor $\pi_i$ by using the genetic algorithm.

\subsubsection{Individual representation}
The pattern chromosome in this algorithm is an vector of n element $X =\{\rho_{\pi_1},\rho_{\pi_2},\ldots,\rho_{\pi_n}\}$. Each element $\rho_{\pi_i}$ is real number in range [0, 1] which represents the ratio of charging time at $\pi_i$ to the total charging time of MC in the charging round. The charging time at $\pi_i$ is then decoded from $\rho_{\pi_i}$ as follows.
\begin{equation}
	\tau_{\pi_i} = T_{charge} \times \frac{\rho_{\pi_i}}{\sum_{i=1}^{n}\rho_{\pi_i}},
\end{equation}

where $T_{charge}$ is the total charging time that is calculated by Eq. \ref{sumTc}.

\subsubsection{Genetic operators}
The initial population consists of $N$ individuals. For each individual $X$, the values of all the genes are assigned by a random number following the uniform distribution in the interval [0, 1].
Two well-known operators, namely Simulated Binary Crossover and Polynomial Mutation \cite{deb2014analysing} are deployed to generate the offspring in each generation. 
Then, $N$ fittest individuals are selected from the current population and the offspring to form the next generation population.

\section{Multitasking approach-based bi-level charging scheme optimization}
\label{sec:propose_2}

\subsection{Motivation of proposed algorithm}

The MLSGA algorithm has overcome limitations of the separated two-phase approach in the work \cite{huong2020genetic} by utilizing the bi-level optimization approach. However, the obtained solution at each iteration of the MLSGA algorithm depends highly on the charging path construction. That means an ineffective path may lead to a sub-optimal result. Therefore, instead of establishing only one good path at the upper level in each iteration to optimize the charging time for the lower level, multiple diverse paths will be formulated at the upper level to explore the search space. The charging paths will be optimized through the evolution process and linked to a lower level to optimize the charging time. As a result, the evolution process is carried out at both levels simultaneously, and the final optimal solution is obtained.
Our second proposed algorithm is constructed based on this basic idea.

In this section, we represent our proposal named MTBCS that adopts the nested evolutionary strategy to assure the diversity of the lower-level search space while retaining a feasible searching time. Specifically, a hybrid of the Genetic Algorithm (GA) and the Local Search (LS) operators is adopted to speed up the charging path optimization at the upper level. Each feasible candidate of the upper level becomes an input for optimizing the charging time at the lower level.
However, instead of optimizing the charging time for all charging paths at the upper level, we divided the charging paths into groups and only chose a representative charging path for each group to identify an optimal charging time at the lower level.
Multiple charging times associated with the chosen charging paths will be optimized simultaneously by the Multitasking Covariance Matrix Adaptation evolution strategy (M-CMAES) based on an explicit knowledge transfer mechanism.

The reason is as follows: 
\begin{itemize}
    \item Genetic Algorithm is strongly capable of identifying promising regions of the search space (exploration), but it often fails or takes a long time to refine the optimal local solution (exploitation), which can be efficiently achieved by a local search method. Therefore, incorporating Local Search into GA may help to obtain a better solution for the upper level.
    \item \label{algo:lower-level} The individuals of the upper level (i.e., the input of the lower level) more or less have certain similarities in terms of the genotypes, while multitasking via knowledge transfer among optimization tasks, has been shown in such researches \cite{huynh2020multifactorial, thang2021adaptive} to be able to enhance the search performance. Moreover, CMA-ES is recognized as a state-of-the-art stochastic algorithm to address black-box continuous optimization problems. Unlike GA, where the individual's evolution is achieved explicitly using predefined crossover and mutation methods, CMA-ES individuals are self-adapting via the Covariance Matrix update and population step size. Because of that, CMA-ES evolution is accomplished implicitly, which has shown to be a more efficient way to improve the quality of the result with faster convergence speed compared to GA in more complex problems like ours since the fixed parameters of GA won't be able to adapt to different changes happen in the evolution process. Thus, a synergy of the multitasking approach and CMA-ES algorithm could establish a robust optimizer for the lower level.
\end{itemize}

The second proposed algorithm is named Multitasking approach-based Bi-level Charging Scheme (MTBCS).

\subsection{MTBCS algorithm scheme}
Algorithm \ref{algo:mfea} represents the pseudo-code of our proposed algorithm MTBCS. After generating and evaluating the initial population (lines 2-3), each iteration of the main loop includes two stages corresponding to two levels of the problem. In the first stages (lines 6-15), the genetic operators and the local search procedure are performed to optimize the charging path. The greedy method proposed in Subsect. \ref{algo:greedy_eval_method} is reused in this stage to evaluate all the generated paths. After that, in the second stage, several paths are selected from the upper level population and the proposed M-CMAES is invoked (lines 16-17) for optimizing the charging time.

\begin{algorithm}[H]
	\label{algo:mfea}
	\caption {\textbf {Multitasking approach-based bi-level charging scheme}}
	\KwIn{A list of sensors ${s_1,s_2,\ldots,s_n}$ , a mobile charger MC.}
	\KwOut{A charging schedule consists of a charging path $\wp^* = (\pi^*_1, \pi^*_2\ldots \pi^*_n)$  and the charging time at each sensor $\mathcal{T}^* = (\tau_{\pi^*_1},\tau_{\pi^*_2},\ldots,\tau_{\pi^*_n})$.}
	\Begin{
		Generate $N$ initial charging paths to form the population $\wp_0$\;
		Evaluate all the charging path in $\wp_0$ \tcp*{using method in Subsect. \ref{algo:greedy_eval_method}}
		$t \longleftarrow 0$\;
		\While{terminate conditions are not satisfied}{
			Offspring $C \longleftarrow \emptyset$\;
			\While{size of $C < N$}{
				Select two parents $p_1$ and $p_2$ from $P_t$ using binary tournament selection\;
				Apply genetic operators on $p_1$ and $p_2$ to produce two offsprings\;
				\For{each offspring $o \in c$}{
					Evaluate offspring $s$\tcp*{using method in Subsect. \ref{algo:greedy_eval_method}}
					\If{$f(o) < f(p_1) \ and \ f(s) < f(p_2)$}{
						localSearch($o$)\tcp*{using local search operator in Subsect. \ref{algo:local_search_operator}}
					}
				} 
				$C \longleftarrow C \bigcup c$\;
			}
			Select $N$ fittest individuals from $P_t \bigcup C$ to form $P_{t+1}$\;
			
			Cluster $P_{t+1}$ into $k$ groups using a k-medoids algorithm\tcp*{see Subsect. \ref{algo:cluster}}
			Optimize charging time for $k$ best charging paths of $k$ groups using M-CMAES algorithm\tcp*{see Subsect. \ref{algo:lower_level}}
			
			$t \longleftarrow t+1$\;
		}
		\textbf{return} $bestSolution$\;
	}
\end{algorithm}	

The next subsection \ref{algo:upper_level} indicates how we employ genetic algorithm and local search at the first stage (upper level). The selecting input for the second stage and details about this stage (lower level) is represented in Subsection \ref{algo:lower_level}.

\subsection{Genetic algorithm for upper level optimization}
\label{algo:upper_level}
\subsubsection{Individual representation}
Each individual presents a charging path for MC, which is a sequence of all sensors in the network. Thus, the permutation encoding is used in this algorithm, where each individual is a string of $n$ natural numbers in range [1, $n$]. An individual is feasible if all the elements are not overlapping with others.

\subsubsection{Genetic operators}
$N$ individuals of the initial population of the upper level are greedily generated by the method in Subsection \ref{algo:mlsga:step1} with the same motivation. In each generation, we employ the Partial Mapped Crossover (PMX) and the Swap Mutation \cite{eiben2003introduction} to produce the offspring. Every time a generated offspring is better than all of its parents, the local search operator proposed in Subsect. \ref{algo:local_search_operator} is performed on that offspring until the local optima is reached. This way, all of the promising regions of the search space that were explored by the genetic operators will be exploited rapidly.

After that, $N$ best individuals are selected from the combination of current population and the generated offspring to form the next generation population.

\subsection{Multitasking covariance matrix adaptation evolution strategy for lower level optimization}
\label{algo:lower_level}
Determining the optimal charging time for a given path is a hard problem, so it is computationally very expensive to optimize the charging time for all generated paths (i.e., perform the lower-level optimization for all members of the upper level population). Therefore, only several paths are selected from the upper level to be the input of the lower level task. In this subsection, we first represent how to perform that selection, and then the proposed M-CMAES  will be presented in detail.

\subsubsection{ K-medoids algorithm for selecting charging paths}
\label{algo:cluster}
In this work, we propose a novel selection mechanism that does not need to perform the lower-level task for every member of the upper level but still covers the whole space of the upper level's population. In particular, we first leverage a k-medoids algorithm to cluster the population of the upper level into $k$ groups based on individuals' similarity. Then, $k$ best individuals of $k$ groups (one best individual for each group) are selected as the input for the lower level.

\begin{figure}
    \centering
    \includegraphics[width=0.3\textwidth]{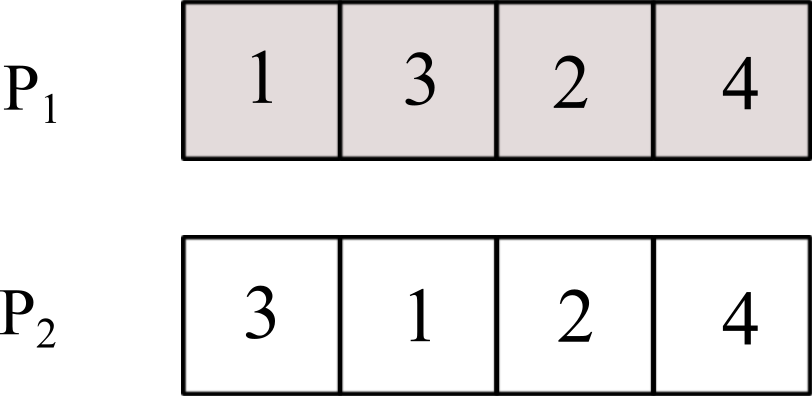}
    \caption{An example of calculating similarity between two charging paths. There are 5 pair of sensors: $\langle1, 2\rangle$, $\langle1, 4\rangle$, $\langle3, 2\rangle$, $\langle3, 4\rangle$, and $\langle2, 4\rangle$ that appear in both $P_1$ and $P_2$. Thus, $\epsilon_{P_1,P_2} = 5$.}
    \label{fig:correlation_example}
\end{figure}

The similarity of individuals in the upper level population is measured based on the correlation in the position of sensors in the decoded paths (phenotypes). 
Let $\epsilon_{P,P'}$ denotes the similarity between two charging paths $\wp$ and $\wp'$. Algorithm \ref{algo:similarity} represents how to calculate the value of $\epsilon_{P,P'}$. Consider all pairs of sensors $\langle\pi_i, \pi_j\rangle$ in $\wp$, ($1\leq i<j \leq n$, without replacement). $\wp$ and $\wp'$ are similar in the pair $\langle\pi_i, \pi_j\rangle$ if $\langle\pi_i, \pi_j\rangle$ also appears in $\wp'$. In that case, $\epsilon_{P,P'}$ is increased by 1. Thus, $\epsilon_{P,P'} = \frac{n(n-1)}{2}$ is the maximum when $\wp$ and $\wp'$ are exactly the same and $\epsilon_{P,P'} = 0$ is the minimum when $\wp$ is an inverse sequence of $\wp'$. Fig. \ref{fig:correlation_example} illustrates an example of algorithm \ref{algo:similarity}.

\begin{algorithm}[H]
	\label{algo:similarity}
	\caption {\textbf {Calculate the similarity between $\wp$ and $\wp'$}}
	\KwIn{Two charging paths $\wp$ and $\wp'$.}
	\KwOut{The similarity value $\epsilon_{P,P'}$.}
	\Begin{
		$\epsilon_{P,P'} \longleftarrow 0$\;
		\For{$i \leftarrow 1$ to $n-1$}{
		    \For{$j \leftarrow i+1$ to $n$}{
		        \If{the pair of sensors $\langle\pi_i, \pi_j\rangle$ in $\wp$ appears in $\wp'$}{
		            $\epsilon_{P,P'} \leftarrow \epsilon_{P,P'} + 1$\;
		        }
		    }
		}
	}
	\textbf{Return $\epsilon_{P,P'}$}\;
\end{algorithm}

After calculating the pairwise similarity between individuals, a k-medoids algorithm, such as PAM is invoked to cluster the population. Due to the limited space, the deployment of PAM algorithm is omitted. Details about this algorithm could be found in \cite{schubert2019faster}.

\subsubsection{ M-CMAES algorithm for the lower level optimization}

In the CMA-ES \cite{auger2012tutorial}, the multivariate normal distribution is built from its population:
\begin{equation}
    \mathcal{N}(m, \sigma^2\mathcal{C}) \sim m + \sigma\mathcal{N}(0, \mathcal{C}),
\end{equation}
where $\mathcal{C}$ is the covariance matrix, $\sigma$ is the step size, and $m$ is the mean vector of the population. During the evolutionary process, CMA-ES selects top $\mu$ fittest individuals from the population of $\lambda$ individuals for updating vector $m$. $\lambda$ offspring are then sampled from the new mean vector $m$, following the distribution $\mathcal{N}(m, \sigma^2\mathcal{C})$.

In our algorithm, optimizing the charging time for each selected path is considered as a single task in the multitask environment. M-CMAES employs $k$ separative populations for $k$ tasks that evolve simultaneously. During the evolutionary process, knowledge transfer explicitly occurs in each task by replacing several individuals of its current population with the best solutions found so far in other tasks. Outlines of M-CMAES are represented in the Algorithm \ref{algo:mcmaes}. 

\begin{algorithm}[H]
	\label{algo:mcmaes}
	\caption {\textbf {M-CMAES}}
	\KwIn{$k$ charging paths that were selected from the upper level population and corresponding $k$ sequence of initial charging time.}
	\KwOut{Optimal charging time for $k$ paths.}
	\Begin{
		Generate $k$ initial populations for $k$ tasks and evaluate all individuals for their associated task only\;
		$t \longleftarrow 0$\;
		\While{terminate conditions are not satisfied}{
			\For{each task $i \leftarrow 1$ to $k$}{
			    Randomly replace $k-1$ individuals in the population of task $i$ by $k-1$ so far best solutions found by the rest $k-1$ tasks\;
			    Apply standard steps of CMA-ES to generate new population for task $i$, i.e., selection, sampling, and distribution updating\;
			    Evaluate all individuals in the new population for task $i$ only\;
			}
		}
	}
\end{algorithm}

The key difference of proposed algorithm from the standard CMA-ES is the benefit of knowledge transfer between tasks. M-CMAES randomly replaces $k-1$ individuals of a task by the best solutions found so far by the rest $k-1$ tasks. By doing that, each task is allowed to inherit the achievement or the good genetic material of other tasks. The search performance of the algorithm is therefore improved, compared to solving each task in a standalone way. It is notable that since the input of all tasks is selected from the population of the upper level, it has certain similarities between tasks, i.e., all the tasks are highly relevant. Thus, the individual replacement can be efficiently performed without the control of random mating probability as in the traditional multifactorial evolutionary algorithm \cite{gupta2015multifactorial}.

It is also worth mentioning that for each task, we leverage the charging time calculated by the greedy method in Subsect. \ref{algo:greedy_eval_method} (Eq. \ref{tau_greedy}) as a good starting point for the evolutionary strategy. The value of $\lambda$ is set as default as $\lambda = \floor[\big]{4 + 3\text{ln}(n)}$, where $n$ is the number of sensors \cite{nishida2018psa}. Thus, the number of tasks in the M-CMAES, i.e., the number of clusters in the k-medoids algorithm (Subsect. \ref{algo:cluster}) is determined by $k = \ceil[\big]{\frac{N}{\lambda}}$, where $N$ is the population size of M-CMAES.

\section{Evaluation Results}
\label{sec:evaluation}
This section evaluates the performances of our two proposed algorithms: Hybrid of \textbf{M}ulti-start \textbf{L}ocal \textbf{S}earch and \textbf{G}enetic \textbf{A}lgorithm (MLSGA) and \textbf{M}ultitasking approach-based \textbf{B}i-level \textbf{C}harging \textbf{S}cheme (MTBCS). Furthermore, we compare our proposals with the most relevant works including an on-demand charging strategy named INMA \cite{zhu2018adaptive}, a periodic charging strategy named GACS \cite{huong2020genetic}, and HPSOGA from \cite{lyu2019periodic}.

INMA sends a charging request to the MC whenever a sensor energy level drops below a predefined threshold. Then, the next to-be-charged node in the charging path will be selected based on their residual energy and Euclidean distance to the current location of the MC. In detail, the set of charging sensor candidates is selected to minimize the number of other requesting nodes that may suffer from energy depletion. 
The GACS algorithm decomposed the task of determining the charging schedule into two sub-problems: finding the optimal charging path and determining the optimal charging time for each sensor. An approximate algorithm using the genetic approach is applied for each sub-problem to find an efficient solution. Then the final charging scheme is constructed using the results of the two sub-problems.
In HPSOGA, periodic charging planning for a mobile WCE with limited traveling energy is proposed. With the optimization objective of maximizing the docking time ratio, this periodic charging planning ensures that the energy of the nodes in the WRSN varies periodically and that nodes perpetually fail to die.

\subsection{Network instances}
 The sensors are deployed in the network with the sensor field size of 500$m$ $\times$ 500$m$ and the base station positioned at the center. A total of 120 network instances are divided into three different types of networks based on the distribution method of the sensors:
	
\begin{itemize}
\item \textbf{Uniform distribution}:  Coordinate $ (x, y) $ of each sensor node is generated by a random function, where the $x_{coordinate}$ and $y_{coordinate}$ are real value in range $ [0, W] $.
\item \textbf{Normal distribution}: The location of sensor nodes is generated following the Gaussian distribution. The sensors generated outside the sensor field will be discarded and regenerated to be inside.
\item \textbf{Grid distribution method}: We divide the network area with dimensions $ W \times W $ into $ 100 $ square cells. Each square cell has a side equal to $ W/10 $. We then randomly generate the location of sensors on each square cell.
\end{itemize}

For each deployed method, 40 network instances are divided into four sets based on their number of sensors: 25, 50, 75, 100, respectively. For each set with these settings, ten different network instances are generated. Each instance is named according to the following format ``\textit{Type\_Num\_Ord}" with the Type corresponding to the distribution type of the network: ``\textit{r}" for Random distribution, ``\textit{n}" for the Normal distribution and, ``\textit{g}" for the Grid distribution; Num is the number of sensors, and Ord is the order of the instance in its set.\\
\subsection{Simulation settings}

\indent We use the MC parameters provided in the works \cite{shi2011renewable,lyu2019periodic}. The average energy consumption rate of each sensor is estimated using the method in \cite{zhu2018adaptive} where the base station collects the information related to the sensor's energy. The detail of the charging model parameters is presented in table \ref{tab:charging_param}. The INMA threshold of sending charging request $e_{thred}$ is set to $0.4 \times$ battery capacity of the sensor. Based on the experiments, all the genetic operators and parameters of algorithms MLSGA, MTBCS, GACS are kept identical as in table \ref{tab:params}.\\

\begin{table}[!htb]
\parbox[t]{.35\linewidth}{
\caption{Charging model parameters}\label{tab:charging_param}
\begin{tabular}{ll}
\hline
\textbf{Parameters} & \textbf{Value} \\ \hline
$E_{MC}$ & 108000 (J)\\ 
U & 5 (J/s) \\ 
$P_M$ & 1 (J/s)\\ 
V & 5 (m/s) \\ 
$e^{max}$ & 10800 (J) \\ 
$e^{min}$ & 540 (J) \\ \hline
\end{tabular}
}
\parbox[t]{.65\linewidth}{
\caption{The proposed algorithms' parameters}\label{tab:params}
\begin{tabular}{ll}
\hline
\multicolumn{1}{l}{\textbf{Parameter}} & \textbf{Value} \\ \hline
Number of   charging path evaluations  & 25000          \\ 
\begin{tabular}[l]{@{}l@{}}Number of charging time evaluations for \\ one charging path\end{tabular} & 25000 \\ 
Population size for optimizing the charging path & 100 \\ 
Population size for optimizing the charging time & 100 \\ 
Crossover rate  & 0.9 \\ 
Mutation rate & 0.05 \\ 
SBX distribution index & 2 \\ 
Polynomial Mutation distribution index & 5 \\ \hline
\end{tabular}}
\end{table}

\indent All experiments are implemented in the JMetalPy framework and conducted on a computer with an Intel Core i7-6700HQ CPU and 8GB of RAM\\

\subsection{Evaluating criteria}
The algorithm efficiency will be evaluated by the dead node ratio of the network calculated by:
	\begin{center}
	 node failure ratio (\%) = $\dfrac{Number of dead nodes}{Number of sensors}\times100$
	 \end{center} 

We focus on the dead node ratio criteria because this is one of the most critical measurements to evaluate the efficiency of a charging scheme in WRSNs.
\subsection{Experimental Results}
To study the effects of the proposed algorithms in solving the \glspl{wrsn} energy depletion problem and experiment with their performances in comparison with existing works, we perform five experiments on the received results:
\begin{itemize}
    \item \textbf{Experiment 1}: Evaluate the effect of our proposed algorithms in comparison with INMA, GACS, and HPSOGA using statistical tests.
    \item \textbf{Experiment 2}: Evaluate the impact of the sensors parameters (number of sensors, average energy consumption rate) on the node failure ratio of the network.
    \item \textbf{Experiment 3}: Evaluate the impact of the MC parameters ( charging rate, MC battery capacity) on the node failure ratio of the network.
    \item \textbf{Experiment 4}: Evaluate the MTBCS convergence trend when applying the greedy and random initialization approaches.
    \item \textbf{Experiment 5}: Evaluate the algorithms running time.
\end{itemize}
\subsubsection{Non-parametric statistic for comparing results of proposed algorithms and existing algorithm}

\indent In recent years, the use of statistical tests to enhance the evaluation process of a new method's performance has become a common technique in computational intelligence \cite{kaswan2018efficient,kumar2020efficient}. Considering the complexity of the charging scheduling problem in \glspl{wrsn}, the requirements of parametric tests (i.e., independence, normality, and homoscedasticity) will not be satisfied. Therefore, to examine the performances of five algorithms MTBCS, MLSGA, GACS, INMA, and HPSOGA, we use Non-parametric tests to analyze the received results.\\
	\indent This study includes three main steps:
	 	\begin{itemize}
	 		\item The first step is to use statistical tests named Friedman and Friedman Align to justify any significant differences among the performances of the algorithms.
	 		\item The second step is to perform multiple comparisons with a control method to highlight the differences in performance of the best algorithm and others.
	 		\item The final step is to perform the median contrast estimation to estimate the magnitudes of the differences between the performance of algorithms.
		\end{itemize}

\begin{table}[h]
\centering
\begin{threeparttable}
\caption{Average Ranking of Algorithms}
\label{tab:statistic}
\begin{tabular}{lll}
\hline
Algorithm       & Friedman  & Aligned Friedman \\ \hline
MTBCS           & 1.3101    & 953.2601         \\
MLSGA           & 2.0863    & 1351.8886        \\
INMA            & 2.7637    & 1719.2142        \\
GACS            & 3.9768    & 2783.6119        \\
HPSOGA          & 4.8630    & 3694.5249        \\ \hline
statistic value & 2745.6590 & 656.6239         \\
p-value         & 0.0000         & 2.2019E-10 \\ \hline     
\end{tabular}
\end{threeparttable}
\end{table}

\indent Results of the Friedman and Align Friedman tests in table \ref{tab:statistic} show that there is a statistically significant difference in the performance of 5 algorithms with $p$-value = $0$ and $p$-value = $2.2e^{-10}$ for Friedman and Aligned Friedman tests, respectively.  However, at this stage, we only know that there are differences somewhere between the related groups, but both the Friedman and Aligned Friedman tests can not pinpoint which groups, in particular, differ from each other.

\begin{table}[h]
\centering
\begin{threeparttable}
\caption{z-values and p-values of the Friedman test (MTBCS is the control algorithm)}
\label{tab:friedman_wcontrol}
\begin{tabularx}{\linewidth}{
@{\extracolsep{\fill}} l*7{l}@{}} \hline Algorithm & z-value & p-value & Holm   & Holland & Rom    & Finner \\ \hline
MLSGA     & 10.0606 & 8.2512E-24  & 0.05   & 0.05    & 0.05   & 0.05   \\
GACS      & 34.5639 & 8.7979E-262 & 0.0166 & 0.0169  & 0.0166 & 0.0253 \\
INMA      & 18.8404 & 3.5203E-79  & 0.025  & 0.0253  & 0.025  & 0.0377 \\
HPSOGA    & 46.0518 & 0.0000           & 0.0125 & 0.0127  & 0.0131 & 0.0127 \\ \hline
\end{tabularx}
\end{threeparttable}
\end{table}

\begin{table}[h]
\centering
\begin{threeparttable}
\caption{z-values and p-values of the Aligned Friedman test (MTBCS is the control algorithm)}
\label{tab:alignedfriedman_wcontrol}
\begin{tabularx}{\linewidth}{
@{\extracolsep{\fill}} l*7{l}@{}}
 \hline
Algorithm & z-value & p-value     & Holm   & Holland & Rom    & Finner \\ \hline
MLSGA     & 6.7372  & 1.6141E-11  & 0.05   & 0.05    & 0.05   & 0.05   \\
GACS      & 30.9349 & 4.0545E-210 & 0.0166 & 0.0169  & 0.0166 & 0.0253 \\
INMA      & 12.9454 & 2.493E-38   & 0.025  & 0.0253  & 0.025  & 0.0377 \\
HPSOGA    & 46.3303 & 0.0000          & 0.0125 & 0.0127  & 0.0131 & 0.0127 \\ \hline
\end{tabularx}
\end{threeparttable}
\end{table}

\indent As a result, additional statistical procedures need to be applied to analyze the performance differences between pairs of algorithms. In more detail, effective post-hoc procedures such as Hom, Holland, Rom and Finner are conducted and MTBCS is considered as control algorithm to demonstrate the differences in performance against the remaining algorithms (MLSGA, INMA, GACS, HPSOGA). \\
\begin{table}[]
\centering
\begin{threeparttable}
\caption{Adjusted p-values for the Friedman test (MTBCS is the control algorithm)}
\label{tab:ajusted_friedman}
\begin{tabularx}{\linewidth}{
@{\extracolsep{\fill}} l*6{l}@{}}
 \hline Algorithm & unadjusted p & pHolm       & pHolland & pRom        & pFinner \\ \hline
MLSGA     & 8.2512E-24   & 8.2512E-24  & 0.0000        & 8.2512E-24  & 0.0000       \\
GACS      & 8.7979E-262  & 2.6393E-261 & 0.0000        & 2.6393E-261 & 0.0000       \\
INMA      & 3.5203E-79   & 7.0407E-79  & 0.0000        & 7.0407E-79  & 0.0000       \\
HPSOGA    & 0.0000            & 0.0000           & 0.0000        & 0.0131      & 0.0000      \\ \hline
\end{tabularx}
\end{threeparttable}
\end{table}

\begin{table}[]
\centering
\begin{threeparttable}
\caption{Adjusted p-values for the Aligned Friedman test (MTBCS is the control algorithm)}
\label{tab:ajusted_alignedfriedman}
\begin{tabular}{llllll}
 \hline 
Algorithm & unadjusted p & pHolm       & pHolland   & pRom        & pFinner    \\ \hline
MLSGA     & 1.6141E-11   & 1.6141E-11  & 1.6141E-11 & 1.6141E-11  & 1.6141E-11 \\
GACS      & 4.0545E-210  & 1.2163E-209 & 0.0000        & 1.2163E-209 & 0.0000        \\
INMA      & 2.4930E-38   & 4.9861E-38  & 0.0000        & 4.9861E-38  & 0.0000        \\
HPSOGA    & 0.0000          & 0.0000         & 0.0000        & 0.0000         & 0.0000  \\ \hline      
\end{tabular}
\end{threeparttable}
\end{table}
\indent Table \ref{tab:friedman_wcontrol} and \ref{tab:alignedfriedman_wcontrol} display the results of Hom, Holland, Rom and Finner post-hoc procedures when comparing MTBCS with other four algorithms. Additionally, Table \ref{tab:ajusted_friedman} and \ref{tab:ajusted_alignedfriedman} show the
adjusted p values for each comparison. We can see that in both the Friedman and Aligned Friedman tests, all procedures reject the null hypothesis with a degree of significance $\alpha = 0.05$ which means that a statistically significant improvement can be obtained by applying the MTBCS algorithm instead of the other four. Therefore, MTBCS is considered as the best algorithm among the five algorithms\\
\indent In order to further demonstrate the  magnitudes of the differences between pairs of algorithms, we also conducted the median based contrast estimation among the algorithms.

\begin{table}[!htp]
\caption{Contrast Estimation}
\label{tab:contrast_estimation}
\begin{tabularx}{\linewidth}{
@{\extracolsep{\fill}} l*6{l}@{}}
\hline
 &MTBCS&MLSGA&INMA&GACS&HPSOGA\\
\hline
MTBCS&0.000&-2.490&-4.440&-10.770&-19.500\\
MLSGA&2.490&0.000&-1.950&-8.280&-17.010\\
INMA&4.440&1.950&0.000&-6.330&-15.060\\
GACS&10.770&8.280&6.330&0.000&-8.730\\
HPSOGA&19.500&17.01&15.060&8.730&0.000\\
\hline

\end{tabularx}
\end{table}
Table \ref{tab:contrast_estimation} shows the contrast estimation among pairs of algorithms. As can be seen, focusing in the rows of the table, specifically at the performance of MTBCS, all its related estimators are negative; that is, it achieves a very low dead node ratio considering median estimators. Considering MLSGA performance, the only positive value in its row belongs to MTBCS, and it is still the best performing algorithm compared to the other four (INMA, GACS, and HPSOGA) in our experimental study.

\subsubsection{Impact of the sensors parameters}
\subsubsubsection{The number of sensors} 

\begin{figure}[H]
    \subfigure[Grid distribution network]{
    \label{fig:sen1_medium}
    \includegraphics[width=0.3\textwidth]{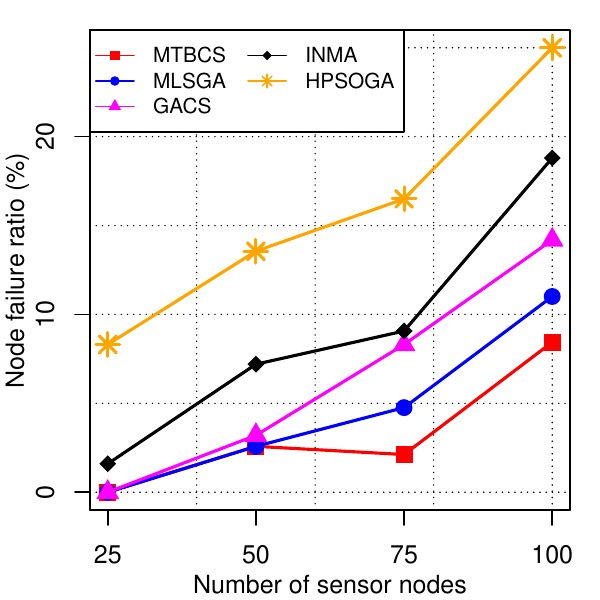}
    \hfill
    }
    \subfigure[Normal distribution network]{
    \includegraphics[width=0.3\textwidth]{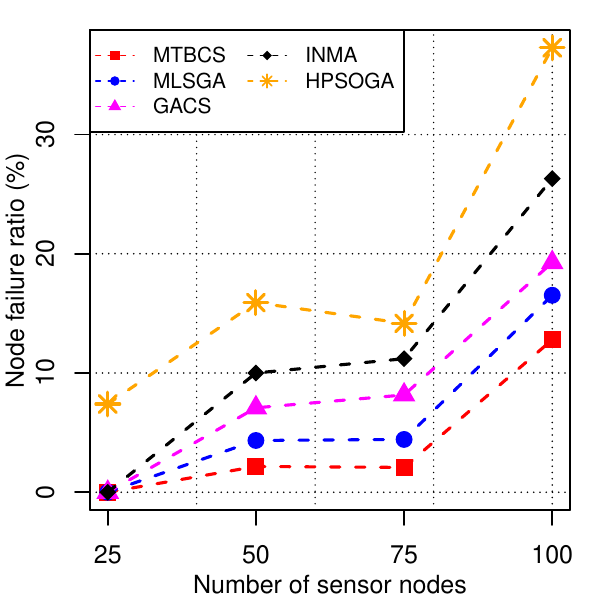}
    \hfill
    }
    \subfigure[Uniform distribution network]{
    \includegraphics[width=0.3\textwidth]{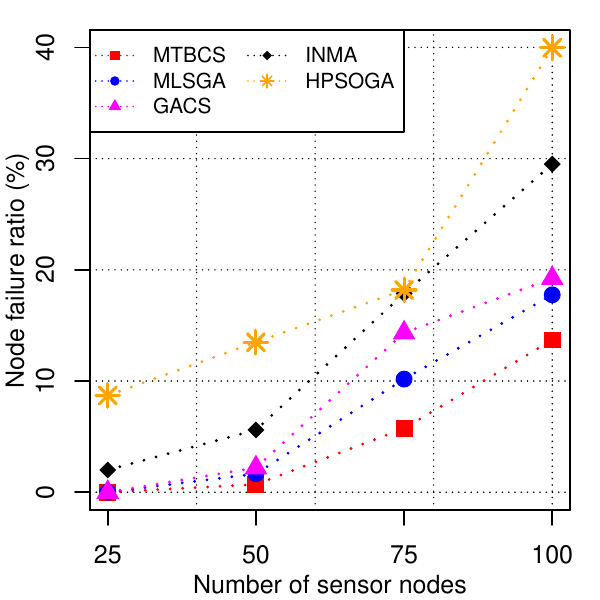}
    \hfill
    }

    \caption{Comparison of the node failure ratio and the number of sensor nodes in different sensor distribution.}
    \label{fig:evalu_sen1}
\end{figure}

\indent Fig \ref{fig:evalu_sen1} illustrates the dead node ratio in three different deployment methods when the number of sensors varies from 25 to 100 nodes. As Fig \ref{fig:evalu_sen1} shows, the dead node ratios of the whole network are proportional to the number of sensors in all three types of network distributions. The reason is that the more sensors there are, the more data packets containing the sensory data and sensor’s residual energy are created. As a result, the sensors have to handle a substantially more enormous workload. Eventually, sensors deplete energy more quickly, which results in the increment of the node failure ratio of the network.\\  

\indent Further comparisons among the results on three distribution methods, we observed that all algorithms perform better when the sensors are deployed following the grid distribution. This phenomenon can be explained by the fact that the grid distribution makes the sensors less concentrated when compared to both the uniform and the normal distribution. Therefore, the workload of all sensors will be reduced since they have to communicate with fewer surrounding sensors. Because of that, the average energy consumption rate of all sensors will reduce, which decreases the node failure ratio of the network.\\

\indent Considering the comparison among algorithms results, we can see that MTBCS and MLSGA significantly outperform INMA, GACS, and HPSOGA in all test cases, which vary in the number of sensors. Specifically, MTBCS reduces the dead node ratio of INMA by 15.81\%, 10.36\%, 13.51\%, and HPSOGA by 26.27\%, 16.56\%, 24.48\%, and GACS by 5.54\% 5.75\% and 6.47\% for u, g and n sets. The MLSGA does not perform as well as the MTBCS, but it still reduces the dead not ratio of INMA by 11.75\%, 11.79\%, 9.78\%, HPSOGA by 22.22\%, 13.99\%, 20.75\%, and GACS by 1.48\%, 3.18\%, and 2.74\% on g, n, and u sets respectively. We can also see that the gap between our two algorithms and INMA tends to widen when the number of sensors increases. This phenomenon could be explained by the fact that INMA only optimizes the sensors in the charging queue but not the rest of the sensors, which will become a problem when the number of sensors is high, leading to a significant number of sensors not being in the queue, hence, the number of not-optimized sensors will also become notable. 
Considering the GACS, even though it has overcome the limitations of the INMA but since the process of determining the charging path and the charging time are separated, the charging path found in the first phase is not guaranteed to benefit the ultimate goal of the algorithm, which is to minimize the dead node percentage of the network. On the contrary, both MTBCS and MLSGA optimize the charging time for multiple charging paths with the only goal to minimize the dead node ratio.
Regarding HPSOGA, the indirect goal to maximize the docking time (time the MC spends recharging itself at the base station) has been ineffective in harsh network conditions where sensor deaths are inevitable. Hence the dead node ratio of HPSOGA increased rapidly with the number of sensors. \\\\
\indent 
 
\subsubsubsection{Impact of average energy consumption rate} 
\begin{figure}[H]
    \subfigure[Grid distribution network]{
    \label{fig:sen7_medium}
    \includegraphics[width=0.3\textwidth]{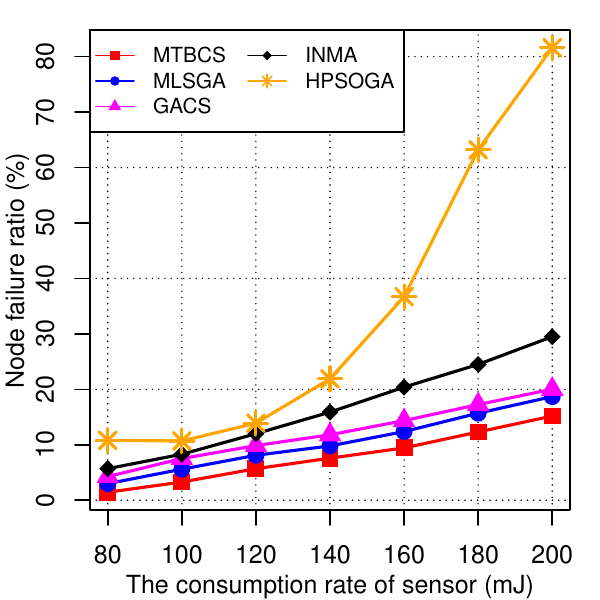}    
    \hfill
    }
    \subfigure[Normal distribution network]{
    \includegraphics[width=0.3\textwidth]{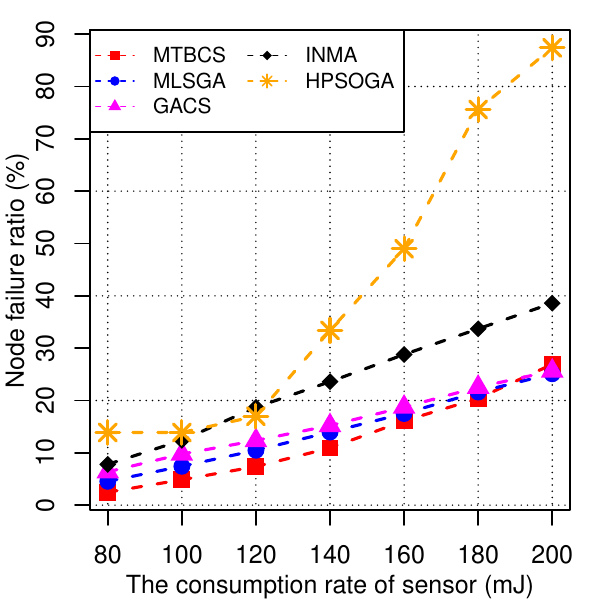}
    \hfill
    }
    \subfigure[Uniform distribution network]{
    \includegraphics[width=0.3\textwidth]{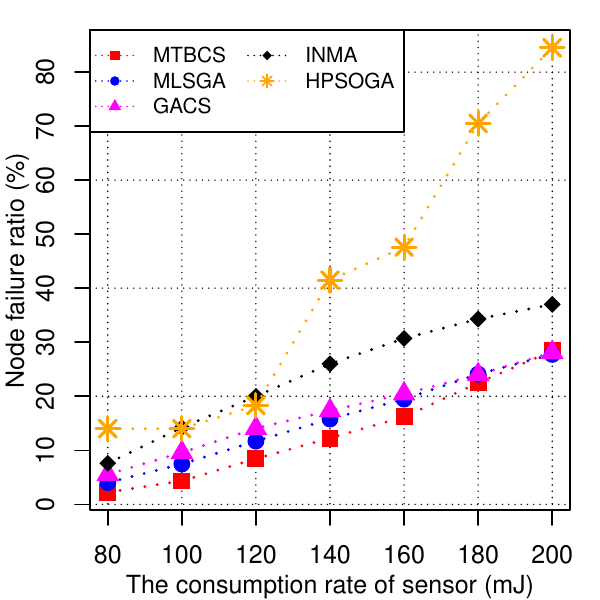}
    \hfill
    }

    \caption{Comparison of node failure ratio and average energy consumption rate for different sensor distributions}
    \label{fig:evalu_sen7}
\end{figure}
\indent Figure \ref{fig:evalu_sen7} demonstrates the node failure ratio of grid network, normal network, and uniform network when the average power consumption rates of sensors are set at various values from $0.8J/s$ to $2J/s$. The line charts show the increasing trend of the dead node ratio when the average energy consumption rate of the sensors increases in all three types of distribution and all four algorithms. Obviously, the higher energy consumption rate results in the shorter sensor's lifetime and the higher energy needed to charge them.\\  

\indent Considering the algorithm's results, we can see that MTBCS and MLSGA significantly outperformed INMA and HPSOGA and showed a better result than GACS. Specifically, on average of MTBCS reduces the dead node ratios of INMA by up to 10.78\%, 8.76\%, 10.65\%, of HPSOGA by 56.09\%, 66.41\%, 60.48\%, and of GACS by 3.53\%, 4.30\%, and 3.08\% with respect to u, g, and n sets. The MLSGA reduces the dead node ratios of INMA up to 8.50\%, 6.14\%, 8.97\%, of HPSOGA by up to 56.75\%, 62.91\%, 62.29\%, and of GACS by 1.26\%, 1.69\%, and 1.42\% on u, n, g sets, respectively.
We can also see that the gap between the periodic charging schemes MTBCS, MLSGA, GACS, and INMA is considerable. As the sensor's consumption rate increases, many charging requests are sent, which increases the number of sensors waiting to be optimized. Whereas the periodic charging schemes, especially our two proposed algorithms, adapt better to the increase of the power consumption rates of the sensors as they optimize all sensors in the network simultaneously.

\subsubsection{Impact of the Mobile Charge  parameters}
\subsubsubsection{Impact of the charging rate}

\begin{figure}[H]
    \subfigure[Grid distribution network]{
    \label{fig:sen2_medium}
    \includegraphics[width=0.3\textwidth]{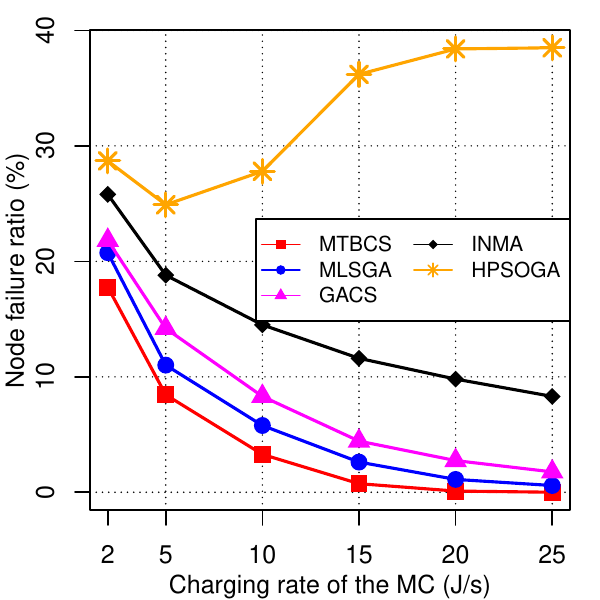}
    \hfill
    }
    \subfigure[Normal distribution network]{
    \includegraphics[width=0.3\textwidth]{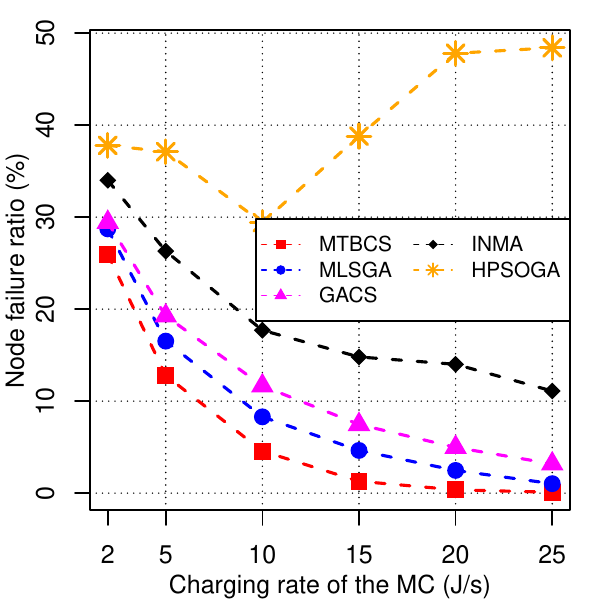}
    \hfill
    }
    \subfigure[Uniform distribution network]{
    \includegraphics[width=0.3\textwidth]{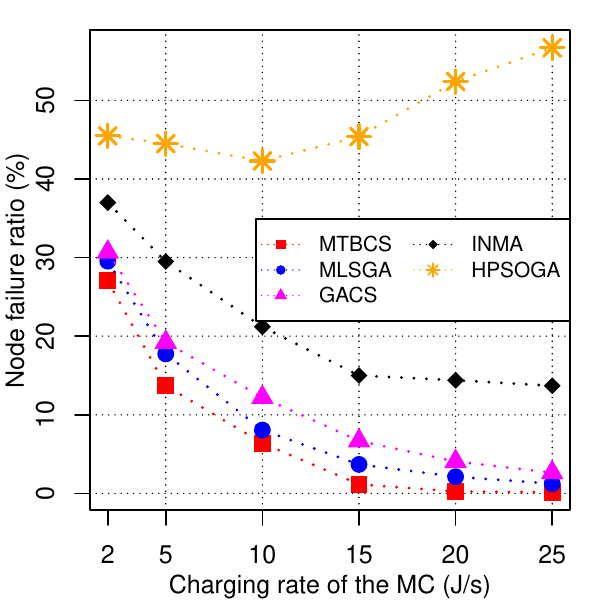}
    \hfill
    }

    \caption{Comparison of the node failure ratio and the charging rate of the Mobile Charger for three distributions}
    \label{fig:evalu_sen2}
\end{figure}
Figure \ref{fig:evalu_sen2} shows the node failure ratio of grid network, normal network, and uniform network when the MC charging power is set at various values from $2J/s$ to $25J/s$. From the three graphs, we can see that the network dead node ratios are inversely proportional to the charging rate of the MC. This is clear because the higher charging rates correspond to less time spent on charging the sensors, making the waiting to be charged time of all sensors lessened. 

\indent Further comparison between algorithms results showed that MTBCS and MLSGA both outperform INMA, GACS, and HPSOGA on all values of the charging ratio of MC. Specifically, MTBCS reduces the dead node ratio rates to 15.81\%, 8.3\%, 11\% when compared to INMA, 56.55\%, 38.5\%, 48.9\% when compared to HPSOGA, and 5.81\%, 1.76\%, 3.08\% when compared to GACS on u, g, n distribution network respectively. MLSGA reduces the dead node ratio rates up to 12.46\%, 7.72\%, 10.07\% when compared to INMA, 55.46\%, 37.92\%, 47.97\% when compared to HPSOGA and 1.4\%, 1.18\%, 2.15\% when compared to GACS.\\
\indent We can also notice that the gaps between INMA and our two algorithms widen when the charging rate increases. For INMA, the charging request is only made when the sensor power drops below a fixed threshold. This approach will not be able to take full advantage of the high MC charging power because the bigger MC charging power should keep all sensors on a larger threshold value. On the other hand, our proposed algorithms can adjust both the charging path and charging time to utilize the high MC charging power.\\
\indent Regarding the GACS, although it also adapts well to the high MC charging rate, because of the separation between its two phases, only the charging time is adapted to the high MC charging rate but not the charging path. Consequently, although it still performs worse than MTBCS and MLSGA, the gap is insignificant even when the charging rate reached $25J/s$.
  
\subsubsubsection{Impact of the MC's battery}
\begin{figure}[H]
    \subfigure[Grid distribution network]{
    \label{fig:sen3_medium}
    \includegraphics[width=0.31\textwidth]{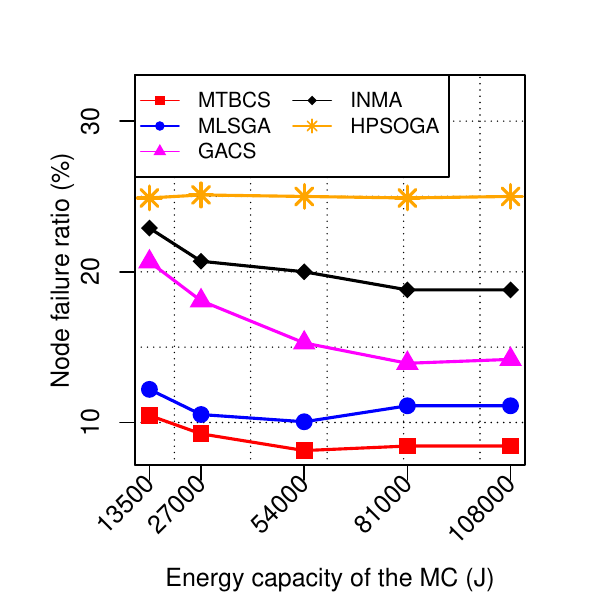}
    \hfill
    }
    \subfigure[Normal distribution network]{
    \includegraphics[width=0.31\textwidth]{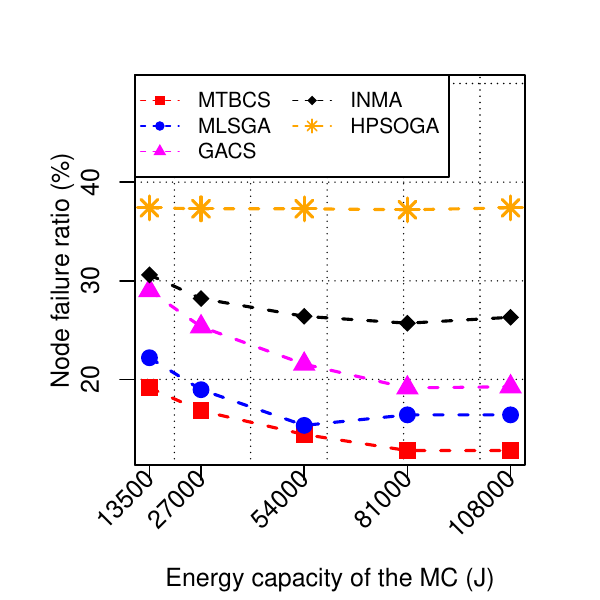}
    \hfill
    }
    \subfigure[Uniform distribution network]{
    \includegraphics[width=0.31\textwidth]{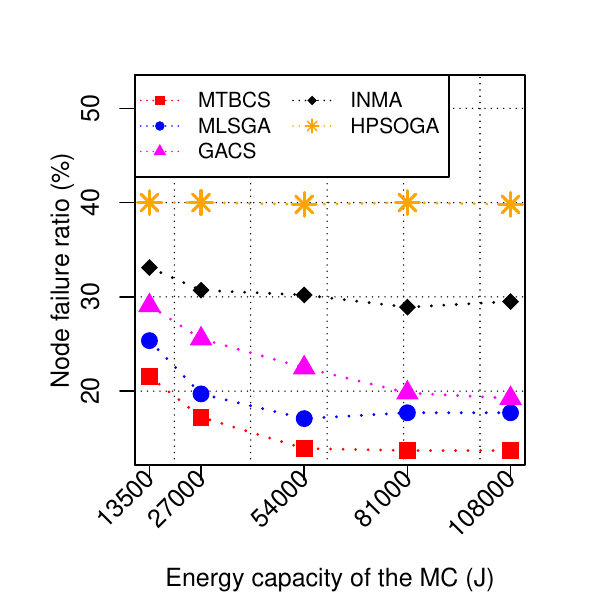}
    \hfill
    }
    \caption{Comparison of the node failure ratio and the battery of Mobile Charger for three distributions}
    \label{fig:evalu_sen3}
\end{figure}
Fig \ref{fig:evalu_sen3} displays the node failure ratio of grid network, normal network, and uniform network when the MC battery capacity is set at various values from $13500J$ to $108000J$. We can see that although the dead node percentage reduces when the MC energy capacity increases, the impact is small and become minuscule as the MC capacity reach a certain value. Specifically, when the MC energy capacity increases from $13500J$ to $2700J$, the dead node ratio only declines 4.33\%, 1.22\%, 2.34\% for the MTBCS algorithm and 5.65\%, 1.67\%, 3.24\% for the MLSGA on u, g and n distribution network, respectively. Furthermore, when the MC capacity increases from $81000J$ to $108000J$, the dead node ratio of MTBCS and HLSGA stop decreasing in all three network distributions.\\
\indent Further examining the results among algorithms, we can also see that our algorithms significantly outperform INMA and GACS on all values of the MC capacity. Specifically, MTBCS reduces the dead node ratio of INMA by 15.81\%, 10.36\%, 13.51\%, HPSOGA by 26.31\%, 16.86\%, 24.61\% and GACS by 5.54\%, 5.75\%, 6.47\%, 24.61\% on u, g and n distribution network respectively. In addition, MLSGA reduces the dead node ratio of INMA by 11.8\%, 7.69\%, 9.89\%, 20.79\%, HPSOGA 22.72\%, 14.95\%, 20.99\%, and GACS by 1.53\%, 3.08\%, 2.85\% with respect to u, g and n distribution.\\ 
\indent The above observations have shown that the MC energy capacity does not play a decisive role in the dead node ratio of the network. This can be explained because when the sensor's parameters and the MC charging power are fixed, the MC only needs a fixed amount of energy to charge and move around the network. Hence, when the MC capacity goes pass this fixed value, it no longer has any impact on the dead node percentage of the network. Because of this reason, the following experiment is conducted to examine the impacts of both the MC charging power and energy capacity on the failure node ratio of the network.
\subsubsubsection{Impact of the charging rate and the MC's battery}
  
\begin{figure}[H]
    \subfigure[Grid distribution network]{
    \label{fig:sen6_medium}
    \includegraphics[width=0.3\textwidth]{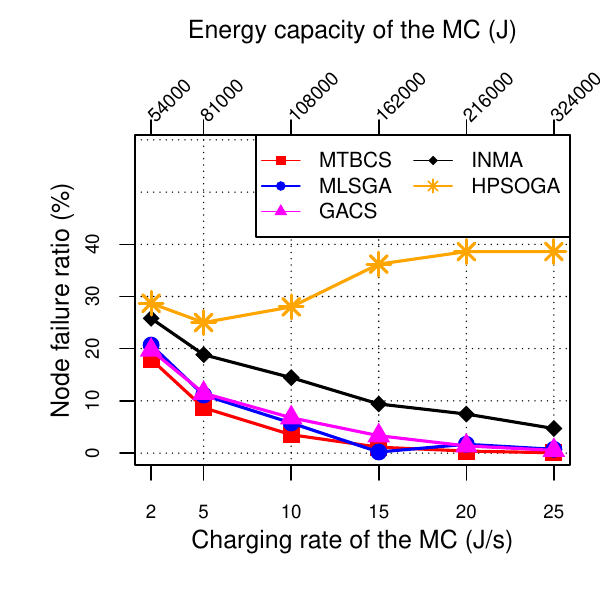}
    \hfill
    }
    \subfigure[Normal distribution network]{
    \includegraphics[width=0.3\textwidth]{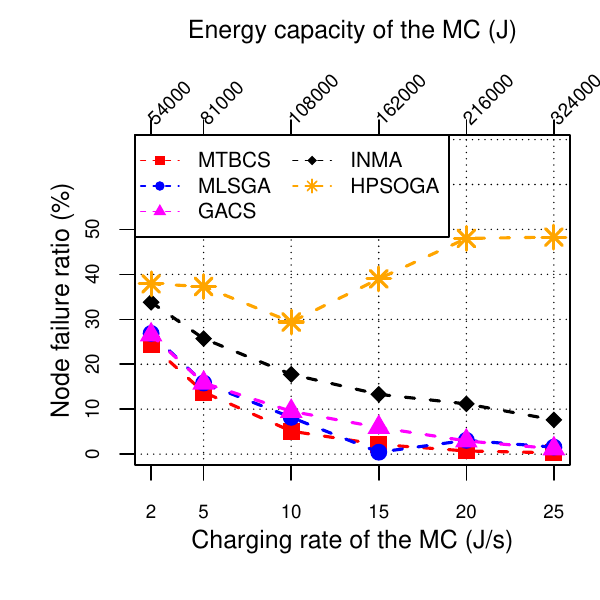}
    \hfill
    }
    \subfigure[Uniform distribution network]{
    \includegraphics[width=0.3\textwidth]{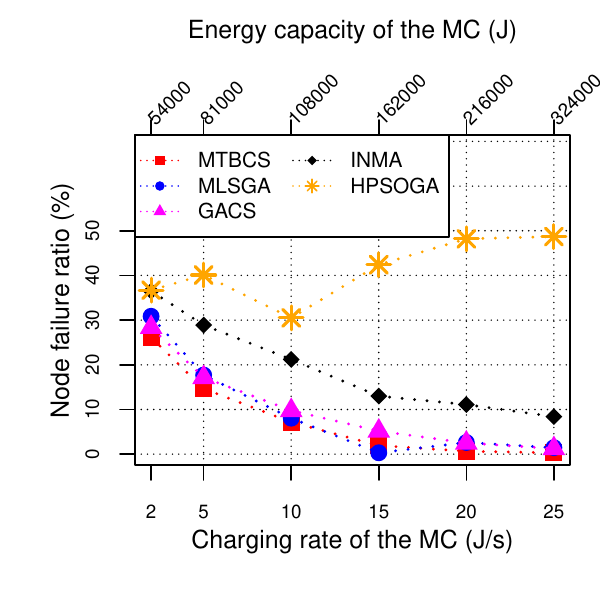}
    \hfill
    }
    \caption{Impact of the charging rate and the MC' battery on the node failure ratio}
    \label{fig:evalu_sen6}
\end{figure}
\indent Figure \ref{fig:evalu_sen6} illustrates the node failure ratio when the MC battery capacity is set at various values from $13500J$ to $108000J$ on a grid network, a normal network, and uniform network. From Figure \ref{fig:evalu_sen6} we can see that the network dead node ratios are inversely proportional to the charging rate and the charging power of the MC. Unlike the only MC energy experiment, the addition of the charging power factor helps substantially reduce the dead node percentage of the network. This could be explained by the fact that it could fully utilize its high energy level when the MC has high charging power. With both high energy and charging power, the MC is capable of charging the sensors faster and replenishing the sensors to a higher level of energy; these stats resulted in a great decrease in the network replenish ratio.\\

\indent Comparing the performances of all five algorithms, we can see that MTBCS outperforms INMA, GACS, and HPSOGA in all experiment varieties while the MLSGA outperforms INMA, HPSOGA and has similar results when compared to GACS. In detail, MTBCS reduces the dead node ratio of INMA up to 11.37\%, 8.16\%, 10.49\%, HPSOGA up to 48.39\%, 38.55\%, 48.00\% , and GACS up to 2.23\%, 1.90\%, 2.59\% on u, g and n distribution network respectively. Furthermore, MLSGA reduces the dead node ratio of INMA by 9.69\%, 6.75\%, 8.91\%, HPSOGA by 47.25\%, 37.89\%, 46.75\%, GACS by 0.54\%, 0.49\%, 1.01\% with respect to u, g, and n network.\\

\subsubsection{Convergence trend of MTBCS}

\begin{figure}[H]
    \includegraphics[width=0.32\textwidth]{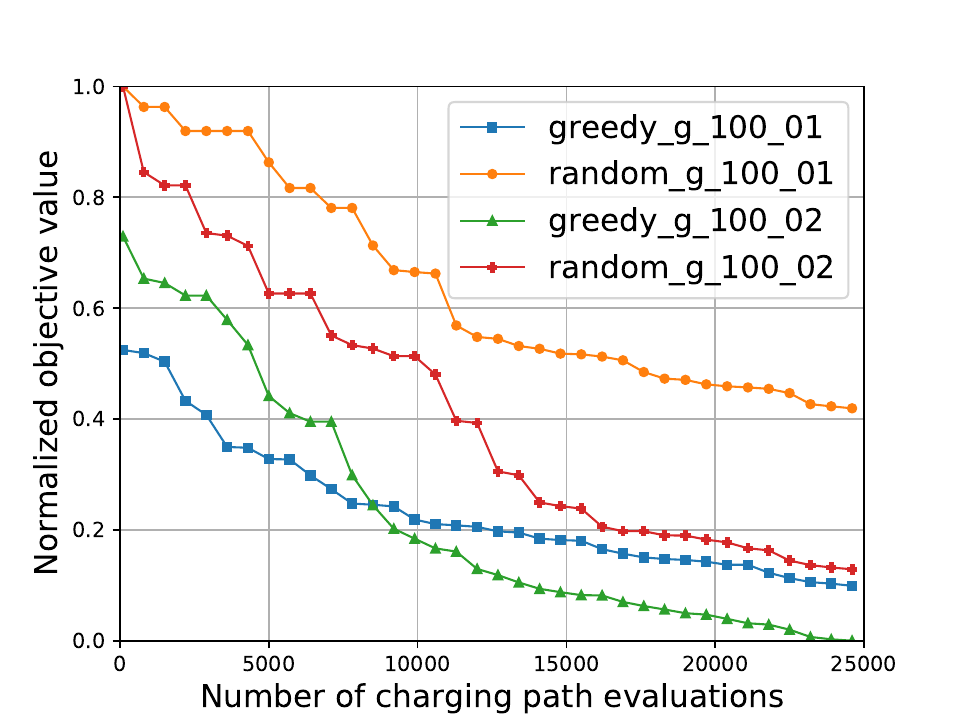}
    \hfill
    \includegraphics[width=0.32\textwidth]{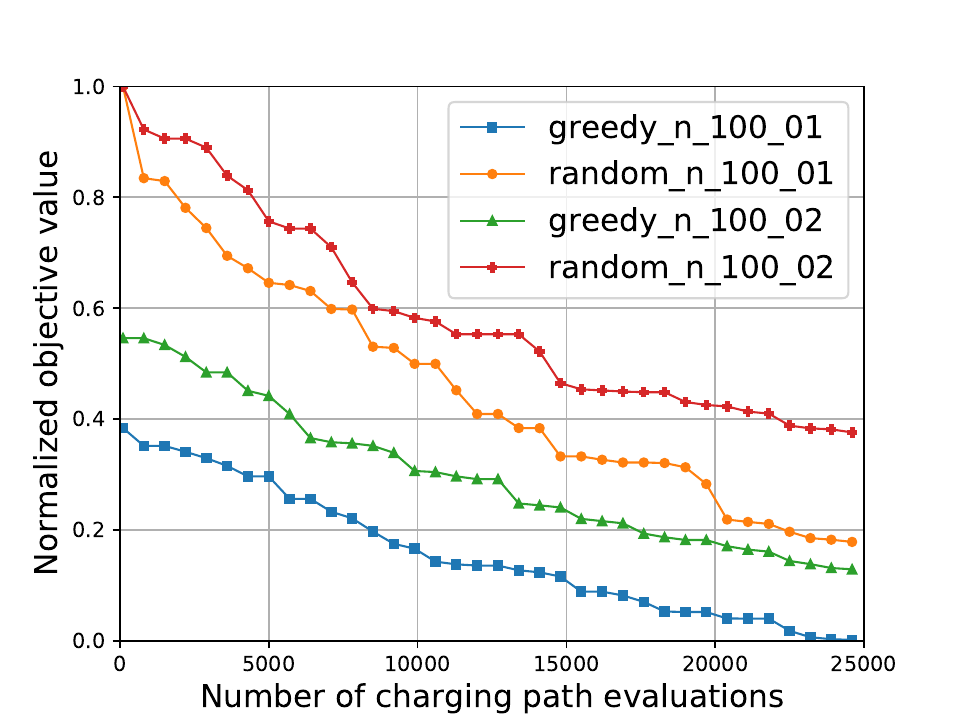}
    \hfill
    \includegraphics[width=0.32\textwidth]{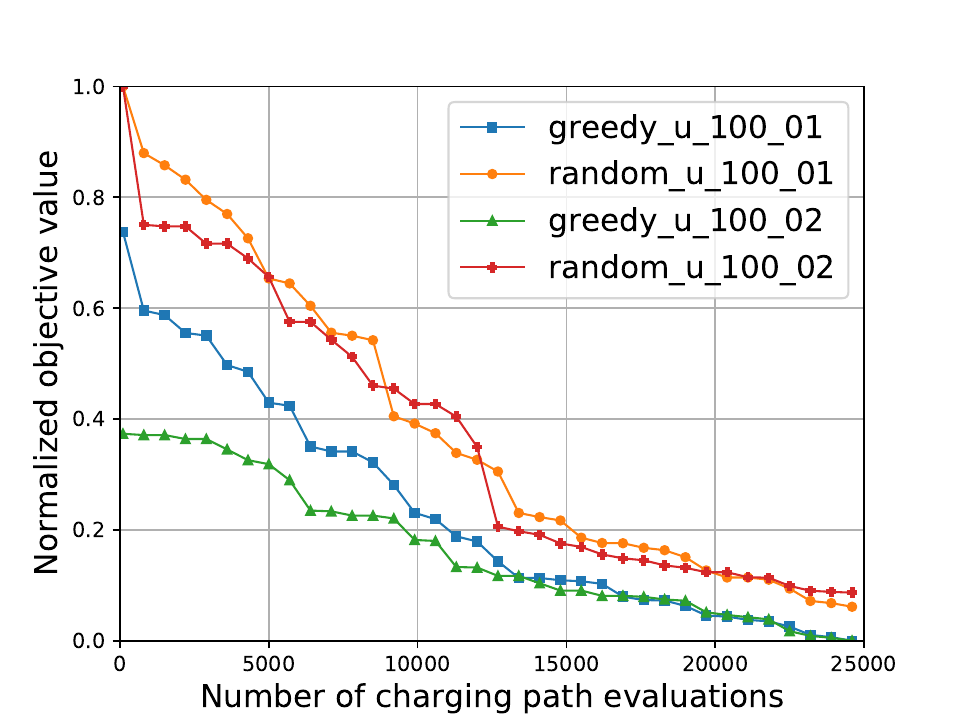}
    \caption{The convergence trend of MTBCS algorithm}
    \label{fig:evalu_convergence}
\end{figure}
\indent Figure \ref{fig:evalu_convergence} illustrates the convergence trends of MTBCS when applying the greedy and the random initialization approach on six instances including two from each type of network: grid network, normal network and uniform network. According to the three line graphs, the greedy initialization method always provide a better starting solution for the searching process. Furthermore, the convergence speed when applying the greedy initialization is also 
consistently higher. Regarding the quality of the final solutions, it can be seen that the MTBCS with the greedy initialization always outperform the random initialization.
\subsubsection{Algorithm's run times analysis}
\begin{table}[H]
\begin{center}
    \caption{Running time comparison between algorithms (\textit{seconds})}
    \label{tab:runtime}
\begin{tabularx}{\linewidth}{@{\extracolsep{\fill}} l*5{c}@{}}
\hline
\textbf{Number of sensors} & \multicolumn{1}{l}{\textbf{GACS}} & \multicolumn{1}{l}{\textbf{MLSGA}} & \multicolumn{1}{l}{\textbf{MTBCS}} &
\multicolumn{1}{l}{\textbf{HPSOGA}}\\ \hline
25  & 0.069 & 0.20  & 1.92 & 12.02 \\ 
50  & 0.18 & 2.05 & 5.44 & 38.56\\ 
75  & 0.36 & 2.34 & 11.52 & 61.66\\ 
100 & 0.48 & 4.08 & 17.4 & 85.70\\ \hline
\end{tabularx}
\end{center}
\end{table}
Table \ref{tab:runtime} shows the detailed running time of MTBCS, MLSGA, GACS, and HPSOGA algorithms. INMA is not included in this evaluation process because it is an on-demand charging algorithm; hence the results of INMA should be obtained at an instance to satisfy the real-time constraints of its charging model. As can be seen, regarding various sensors, GACS is always the fastest algorithm among the three, and HPSOGA is the slowest. MTBCS and MLSGA are the second and third slowest algorithms among the four. This trend can be explained by the fact that GACS only optimizes charging time for the best charging path obtained by the first phase, whereas the charging time optimization is performed once every iteration in MLSGA and $k times$ per iteration in MTBCS. However, considering the proposed algorithms' performance improvements to GACS combined with the static nature of the periodic charging approach, the running time is still acceptable, and the trade-off is worthy. On the other hand, HPSOGA is the slowest algorithm while having the worst result because of its indirect goal to maximize the docking time of the MC.

\section{Conclusions}
\label{sec:conclusion}
In this paper, we study the problem of energy depletion avoidance in \gls{wrsn}. To achieve that ultimate goal, we focus on minimizing the number of dead sensor nodes after the charging process based on an optimized bi-level charging approach where the charging path and charging time are simultaneously solved. Since the search space is enormous and complex, we proposed two meta-heuristic algorithms with two novel search strategies to handle the investigated problem. The first algorithm MLSGA starts from multiple points to explore the search space and then exploits the feasible space by genetic algorithm. The second algorithm, MTBCS, leverages the superiority of the multitasking approach and covariance adaptation evolutionary strategy to optimize charging time at the lower level. Finally, we extensively performed experiments in network scenarios and
the experimental results have demonstrated that our charging algorithms significantly reduce the number of dead nodes compared to the benchmark. 

\section*{Acknowledgment}
\label{sec: ackowledgement}
Funding: This research is funded by Vietnam National Foundation for Science and Technology Development
(NAFOSTED) under grant number 102.01-2019.304.
  
\bibliography{cite}

\end{document}